\documentclass[preprint,12pt,authoryear]{elsarticle}

\usepackage{amssymb}
\usepackage{times}
\usepackage{soul}
\usepackage{url}
\usepackage{booktabs}
\usepackage[hidelinks]{hyperref}
\usepackage[utf8]{inputenc}
\usepackage[small]{caption}
\usepackage{graphicx}
\usepackage{array,multirow}
\usepackage{arydshln}
\usepackage{amsmath}
\usepackage{adjustbox}
\usepackage{booktabs}
\usepackage{algorithm}
\usepackage{algorithmic}
\usepackage[switch]{lineno}
\usepackage{natbib}
\usepackage{subfig}
\usepackage{graphicx}
\setcitestyle{numbers,square}
\begin{document}

\begin{frontmatter}



\title{RF-GNN: Random Forest Boosted Graph Neural Network for Social Bot Detection}

\author[label1]{Shuhao Shi}
\author[label1]{Kai Qiao}
\author[label1]{Jie Yang}
\author[label1]{Baojie Song}
\author[label1]{Jian Chen}
\author[label1]{Bin Yan*}
\ead{ybspace@hotmail.com}

\address[label1]{organization={Henan Key Laboratory of Imaging and Intelligence Processing, PLA strategy support force information engineering university},%Department and Organization
            addressline={45000},
            city={Zhengzhou},
            state={Henan},
            country={China}}

\begin{abstract}
The presence of a large number of bots on social media leads to adverse effects. Although Random forest algorithm is widely used in bot detection and can significantly enhance the performance of weak classifiers, it cannot utilize the interaction between accounts. This paper proposes a Random Forest boosted Graph Neural Network for social bot detection, called RF-GNN, which employs graph neural networks (GNNs) as the base classifiers to construct a random forest, effectively combining the advantages of ensemble learning and GNNs to improve the accuracy and robustness of the model. Specifically, different subgraphs are constructed as different training sets through node sampling, feature selection, and edge dropout. Then, GNN base classifiers are trained using various subgraphs, and the remaining features are used for training Fully Connected Netural Network (FCN). The outputs of GNN and FCN are aligned in each branch. Finally, the outputs of all branches are aggregated to produce the final result. Moreover, RF-GNN is compatible with various widely-used GNNs for node classification. Extensive experimental results demonstrate that the proposed method obtains better performance than other state-of-the-art methods.
\end{abstract}


\end{frontmatter}


\section{Introduction}
\label{sec:Introduction}
Social media have become an indispensable part of people's daily lives. However, the existence of automated accounts, also known as social bots, has brought many problems to social media. These bots have been employed to disseminate false information, manipulate elections, and deceive users, resulting in negative societal consequences~\cite{article01,article02,article03}. Effectively detecting bots on social media plays an important role in protecting user interests and ensuring stable platform operation. Therefore, the accurate detection of bots on social media platforms is becoming increasingly crucial.
Random Forest (RF)~\cite{article04} is a classical algorithm of ensemble learning that can significantly improve the performance of the base classifier, Decision Tree (DT)~\cite{article05}. Specifically, $S$ sub-training sets are generated by randomly selecting n samples with replacement from the original training set of $N$ samples $S$ times. Then, $m$ features are selected from the $M$-dimensional features of each sub-training set, and S base classifiers are trained using different sub-training sets. The final classification result is determined by the voting of the base classifiers. Due to its excellent performance, RF has been widely applied in various competitions, such as data mining and financial risk detection, and is also frequently used in social bot detection. \cite{article06} made use of RF classifer to evaluate and detect social bots by creating a system called BotOrNot. \cite{article07} conducted a study analyzing the impact of political bots on the 2018 Swedish general elections. The study evaluated several algorithms, including AdaBoost~\cite{article08}, Support Vector Machine (SVM)~\cite{article08}, and RF, and found that RF outperformed the other algorithms, achieving an accuracy of 0.957. \cite{article10} conducted an evaluation of several classifiers, including RF, Adaboost, SVM, and K-Nearest Neighbors (KNN)~\cite{article11}, for bot detection. After analyzing the performance of these classifiers, it was determined that blending using RF produced the best results.
Although RF algorithm performs well in social bot detection, new camoulage and adversarial techniques evolve to maintain threats and escape from perception~\cite{article12}. Social bots can simulate genuine users through complex strategies to evade feature-based detection methods~\cite{article01}. Rarely consider the connection between users~\cite{article12}, making it challenging to ensure detection accuracy.
Graph neural networks (GNNs)~\cite{article17,article18,article19} are emerging deep learning algorithms that effectively utilizes the relationship between nodes in graph structures. They have been widely applied in fields such as social networks and recommendation systems. Recently, GNNs have been applied in social bot detection, utilizing user relationships effectively during detection compared to traditional machine learning algorithms. GNN-based approaches~\cite{article13,article15,article16} formulate the detection processing as a node classification problem. \cite{article13} was the first to attempt to use graph convolutional neural networks~\cite{article13} in detecting bots, effectively utilizing the graph structure and relationships of Twitter accounts. Recent work has focused on studying multiple relationships in social graphs, \cite{article13} introduced Relational Graph Convolutional Networks (RGCN)~\cite{article13} into Twitter social bot detection, which can utilize multiple social relationships between accounts. \cite{article16} proposes a graph learning data augmentation method to alleviate the negative impact of class imbalance on models in bot detection.
To effectively leverage the advantages of ensemble learning and GNNs, we have designed a Random Forest boosted Graph Neural Network for soical bot detection, called RF-GNN. Specifically, RF-GNN us GNNs as the base classifiers of the RF algorithm. To construct diverse sub-training sets effectively, we construct subgraphs using node sampling, feature selection, and edge dropout as the training sets for different GNN base classifiers.
We use the remaining features after feature selection as the input for the Fully Connected Neural Network (FCN), aligning the outputs of the GNN base classifiers. As a result, each base classifier is enabled to be insensitive to the sub-training set, increasing the robustness of RF-GNN. Furthermore, the design of the aligning mechanism can effectively utilize the discarded features of the GNN base classifiers. Our proposed RF-GNN framework can be applied to various GNN models and significantly improves the accuracy and robustness of GNN. The main contributions of this work are summarized as follows.
\begin{itemize}
\item We proposed a novel framework is that combine random forest algorithm with GNNs, which is the first of its kind. The framework effectively utilizes the ability of GNNs to leverage relationships and the advantages of ensemble learning to improve the performance and robustness of model.
\item We propose an alignment mechanism that further enhances the performance and robustness of the GNNs ensemble model by effectively utilizing the remaining features after feature selection.
\item Our proposed framework is flexible and can be utilized with various widely-used backbones. Extensive experimental results demonstrate that our proposed framework significantly improves the performance of GNNs on different social bot detection benchmark datasets.
\end{itemize}

\section{Preliminaries}
\label{sec:Preliminaries}
In this section, we define some notations used throughout this paper. Let $\mathcal{G}=(\mathcal{V}, \mathcal{E})$ represent the graph, where $\mathcal{V}$ is the set of vertices $|\mathcal{V}|=\left\{v_{1}, \cdots, v_{N}\right\}$ with $|\mathcal{V}|=N$ and $\mathcal{E}$ is the set of edges. In this paper, we consider multi-relationship social networks graphs. The adjacency matrix is defined as $A^{k} \in\{0,1\}^{N \times N}$, $1 \leq k \leq K$. $K$ represents the total number of edge types, and $\boldsymbol{A}_{i, j}^{k}=1$ if and only if $\left(v_{i}, v_{j}\right) \in \mathcal{E}^{k}$, $\mathcal{E}^{k} \subseteq \mathcal{E}$. Let $\mathcal{N}_{i}$ denotes the neighborhood of node $v_i$. The feature matrix is denoted as $\boldsymbol{X} \in \mathbb{R}^{N \times M}$, where each node $v$ is associated with a $M$ dimensional feature vector $\boldsymbol{X}_{v}$.

\noindent \textbf{Graph Neural Networks}
GNNs~\cite{article17,article18,article19} could directly operate on non-Euclidean graph data, which generates node-level feature representation through message passing mechanism. Specifically, the feature representation of nodes at the $l$ layer is obtained through the aggregation of 1-hop neighborhood at the $l-1$ layer. The $l$-th layer of the GNN message-passing scheme is:

\begin{equation}
\label{eq:GNN}
\mathbf{h}_{v}^{(l)}=\operatorname{COM}\left(\mathbf{h}_{u}^{(l-1)}, \operatorname{AGG}\left(\mathbf{h}_{u}^{(l-1)}, u \in \mathcal{N}_{v}\right)\right)
\end{equation}

where $\operatorname{COM(\cdot)}$ and $\operatorname{AGG(\cdot)}$ denotes COMBINE and AGGREGATE functions respectively, $\mathbf{h}_{v}^{(l)}$ is the representation vector of node $v$ in the $l$-th layer. Specifically, $\boldsymbol{h}_{v}^{(0)}=\boldsymbol{X}_{v}$.

\section{The Proposed Method}
\subsection{Motivation}
\label{sec:Motivation}
The base classifier of RF is DT, which has several disadvantages compared to neural networks. DT is prone to over-fitting on training data, especially when the tree depth is large. They have weak modeling ability for complex nonlinear relationships, and may require more feature engineering to extract more features. DT is also sensitive to noise in the data, leading to weak model robustness. Using graph neural networks as the base classifier of random forest can effectively solve these problems and make effective use of the relationships between accounts. Compared to traditional ensemble learning, introducing relationships between accounts can improve the performance of the model. Compared to previous GNNs, using bagging improves the performance and robustness of the model.

\begin{figure}[h]
   \centering
   \begin{minipage}[b]{0.965\textwidth}
    \includegraphics[width=\textwidth]{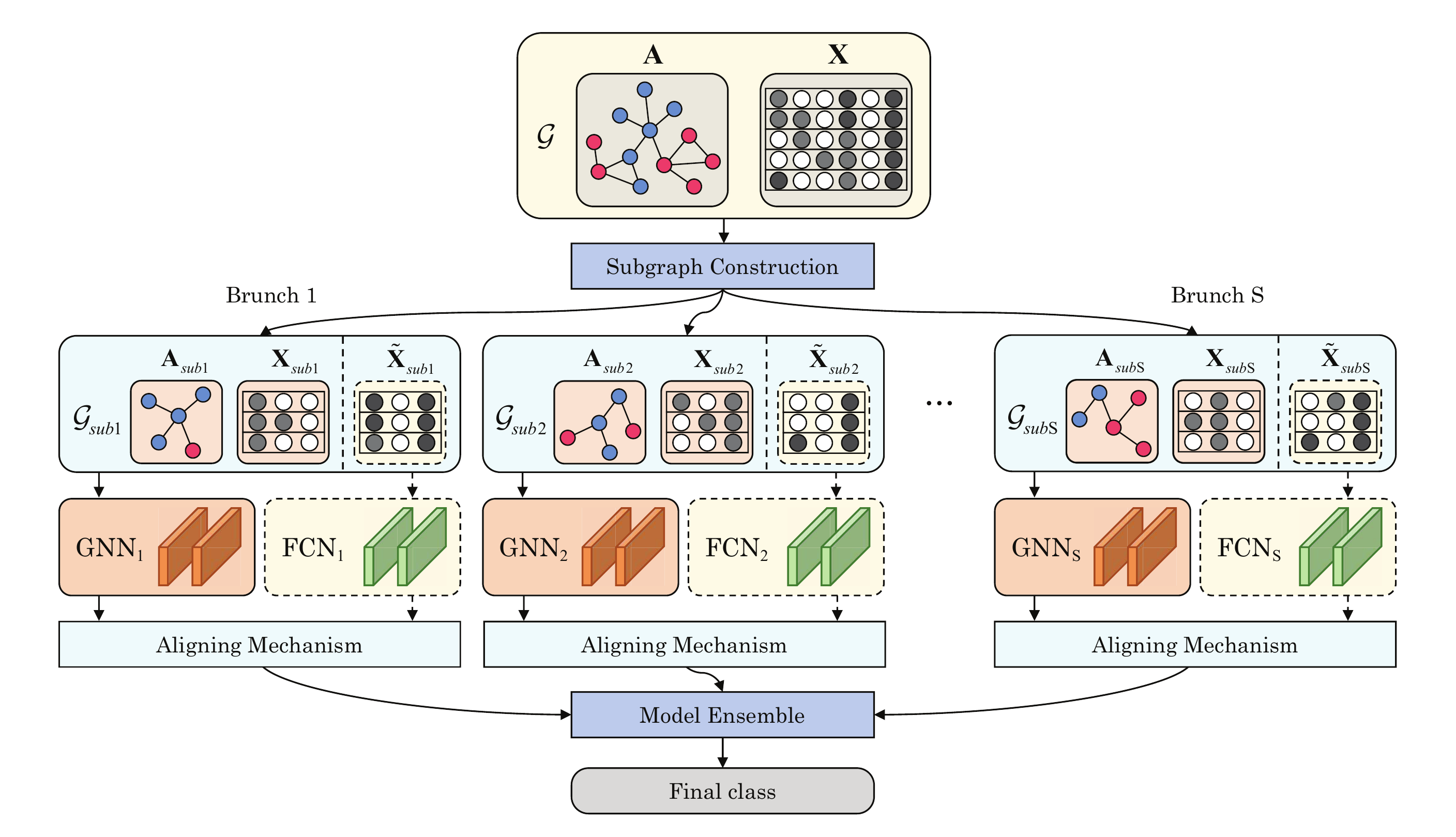}
  \end{minipage}
  \caption{The proposed RF-GNN framework.}
  \label{fig:framework}
\end{figure}

Figure \ref{fig:framework} shows the overall structure of RF-GNN, which consists of the subgraph construction module, the alignment mechanism, and model ensemble module.

\subsection{Subgraph construction}
\label{sec:Subgraph_construction}
The purpose of subgraph construction is to obtain $S$ subgraphs, which are used as training data for $S$ GNN base classifiers. In order to ensure the diversity of the sub-training sets, three methods are used to construct the subgraphs, including node sampling, feature selection, and edge dropping.

\noindent \textbf{Nodes sampling}
Due to the unique characteristics of graph data, sample sampling is no longer done by randomly selecting nodes with replacement, as is the case with the random forest algorithm. In nodes sampling, a certain portion of vertices without replacement along with their connections are randomly selected. The probability of keeping nodes, denoted by $\alpha$, follows an \textit{i.i.d.} uniform distribution.

\noindent \textbf{Feature selection}
Based on the idea of selecting a subset of feature dimensions as new features in the RF algorithm, $\beta$ proportion of feature dimensions are randomly selected from the $M$-dimensional feature vector of subgraph $\mathcal{G}_{sub\text{i}}$ to form $\mathbf{X}_{sub\text{i}}$, which is used as the feature matrix for the input of the $i$-th GNN base classifier.

\noindent \textbf{Edge dropping}
To further increase the differences between subgraphs, we randomly remove a portion of edges in the subgraph $\mathcal{G}_{sub\text{i}}$. The proportion of edges to be dropped is $1-\gamma$, which follows a normal distribution. In other words, $\gamma$ proportion of edges are retained.

\subsection{Aligning mechanism}
\label{sec:Aligning_mechanism}

\begin{figure}[ht]
   \centering
   \begin{minipage}[b]{0.785\textwidth}
    \includegraphics[width=\textwidth]{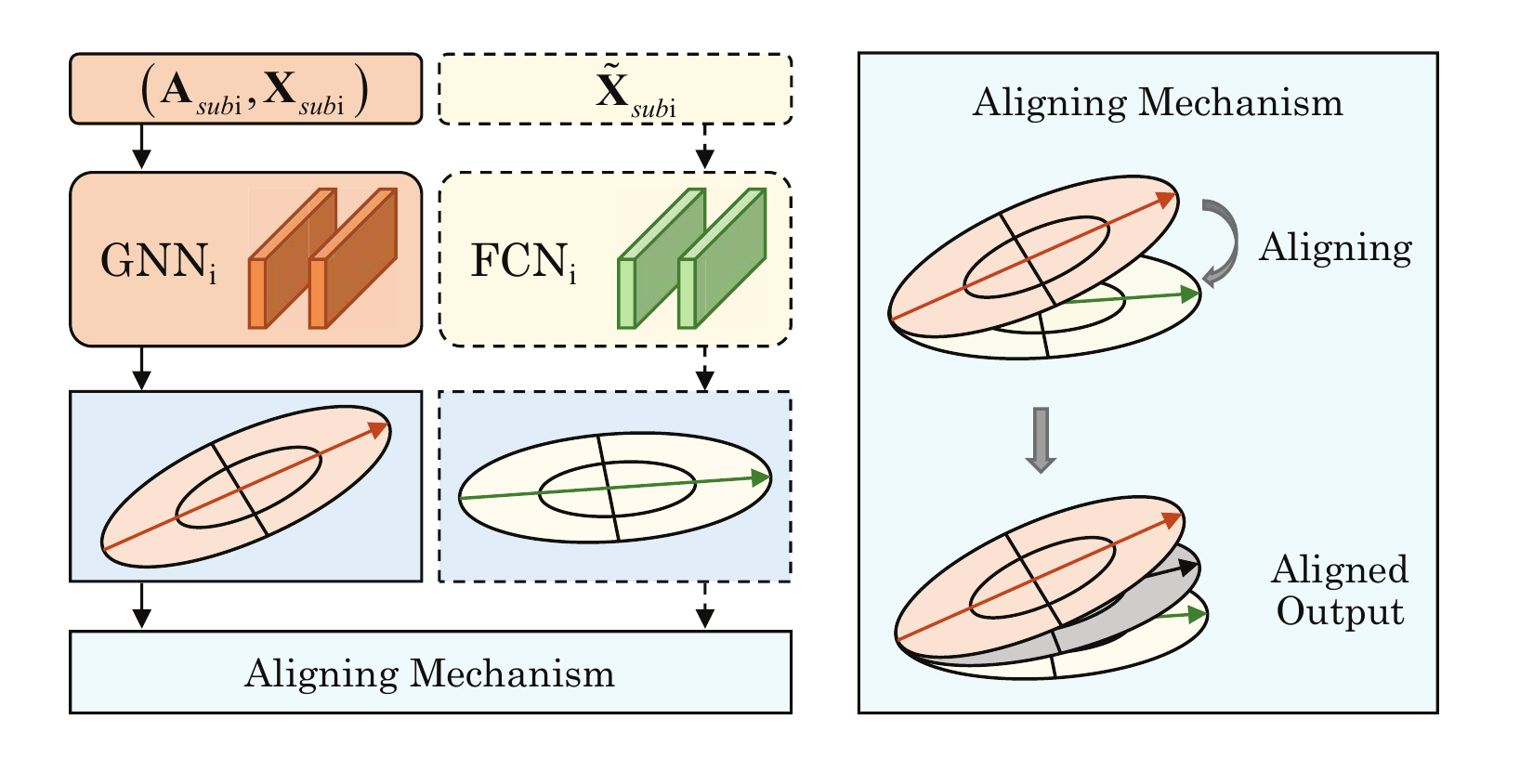}
  \end{minipage}
  \caption{Aligning mechanism in the $i$-brunch of RF-GNN.}
  \label{fig:aligning}
\end{figure}

$S$ subgraphs are constructed for training $S$ different branches. In the $i$-th branch, as shown in Figure \ref{fig:aligning}, only partial dimensional features $\mathbf{X}_{sub\text{i}}$ are used for the training of the base classifier $G_{\theta_{i}}$. The remaining features $\tilde{\mathbf{X}}_{sub\text{i}}$ after feature selection are used to train the fully connected neural network (FCN) $F_{\theta_{i}}$. The output of the GNN-based classifier and the output of the FCN are aligned through the Hadamard product. Through this alignment method, the output between the GNN and FCN is facilitated to enhance the performance and stability of the model:

\begin{equation}
\label{eq:aligning}
\mathbf{Z}_{i}=G_{\theta_{i}}\left(\mathcal{E}_{i}, \mathbf{X}_{s u b i}\right) \odot F_{\theta_{i}}\left(\tilde{\mathbf{X}}_{s u b i}\right).
\end{equation}

We use the output embedding $\mathbf{Z}_{i}$ in Eq. \ref{eq:aligning} for semi-supervised classification with a linear transformation and a softmax function. Denote the class predictions of $i$-th brunch as $\hat{\mathbf{Y}}_{i} \in \mathbb{R}^{N \times C}$, then the $\hat{\mathbf{Y}}$ can be calculated in the following way:

\begin{equation}
\label{eq:predictions}
\hat{\mathbf{Y}}_{i}=\operatorname{softmax}\left(\mathbf{W} \cdot \mathbf{Z}_{i}+\mathbf{b}\right),
\end{equation}

where $\mathbf{W}$ and $\mathbf{b}$ are learnable parameters, softmax is actually a normalizer across all classes. Suppose the training set is $V_L$, for each $v_{n} \in V_{L}$ the real label is $\mathbf{y}_{\text {in}}$ and the predicted label is $\tilde{\mathbf{y}}_{in}$. In this paper, we employs cross-entropy loss to measure the supervised loss between the real and predicted labels. The loss function of the $i$-th branch is as follows:

\begin{equation}
\label{eq:loss}
\mathcal{L}_{i}=-\sum_{v_{n} \in \mathrm{V}_{L}} \operatorname{loss}\left(\mathbf{y}_{in}, \tilde{\mathbf{y}}_{in}\right).
\end{equation}

\subsection{Model ensemble}
\label{sec:Model_ensemble}
The base classifiers of $S$ branches can be trained in parallel. After training the $S$ base classifiers, the output is aggregated to obtain the final classification results:

\begin{equation}
\label{eq:ensemble}
\hat{\mathbf{Y}}=\sum_{i=1}^{S} \hat{\mathbf{Y}}_{i}.
\end{equation}

The learning algorithm is summarized in Algorithm 1.

\begin{algorithm}[ht]
%    \footnotesize
	\caption{RF-GNN Training Algorithm.}
	\label{alg:training}
	{
	\begin{algorithmic}[1]
			\STATE {\bfseries Input:} Graph $\mathrm{G}$, Node representation $\boldsymbol{X} \in \mathbb{R}^{N \times M}$, Node sampling probability $\alpha$, Feature selecting probability $\beta$, Edge keeping probability $\gamma$, GNN base classifiers, $G_{\theta_{1}}, G_{\theta_{2}}, \ldots, G_{\theta_{S}}$, FCN models $F_{\theta_{1}}, F_{\theta_{2}}, \ldots, F_{\theta_{S}}$
			\STATE {\bfseries Output:} Predicted node class $\hat{\mathbf{Y}}$
			\STATE {Initialize the classifier parameter $\theta$}
			\FOR {$i = 1, 2, \ldots, S$}
    		  \STATE {Obtain $\mathcal{G}_{i}\left(\mathcal{V}_{i}, \mathcal{E}_{i}\right)$ by nodes sampling on $\mathrm{G}$ with probability $\alpha$}
              \STATE {Obtain $\mathbf{X}_{sub \mathrm{i}}$ by feature selecting on $\mathbf{X}$ with probability $\beta$, the rest feature denote as $\tilde{\mathbf{X}}_{sub\mathrm{i}}$}
              \STATE {Drop edge on $\mathcal{E}_{i}$ with probability $1-\gamma$}
              \STATE {Training GNN classifier $G_{\theta_{i}}$ using $\mathcal{E}_{i}$ and $\mathbf{X}_{sub \mathrm{i}}$}
        	  \STATE {Training FCN model $F_{\theta_{i}}$ using $\tilde{\mathbf{X}}_{sub\mathrm{i}}$}
              \STATE {Align the output of $\mathcal{E}_{i}$ and $F_{\theta_{i}}$ via Eq. \ref{eq:aligning}}
              \STATE {Update parameters of $G_{\theta_{i}}$ and $F_{\theta_{i}}$ by applying gradient descent to maximize Eq. \ref{eq:loss}}
			\ENDFOR
			\STATE {Ensemble output prediction $\hat{\mathbf{Y}}_{i}$ via Eq. \ref{eq:ensemble}}
	\end{algorithmic}
	}
\end{algorithm}

\section{Experiment setup}
\subsection{Dataset}
We evaluate Twitter bot detection models on three datasets that have graph structures: Cresci-15~\cite{article03}, Twibot-20~\cite{article35}, and MGTAB~\cite{article12}. The detailed description of these datasets is as follows:

\begin{itemize}
\item Cresci-15 is a dataset consisting of 5,301 users who have been labeled as either genuine or automated accounts. The dataset provides information on the follower and friend relationships between these users.
\item Twibot-20 is a dataset containing 229,580 users and 227,979 edges, of which 11,826 accounts have been labeled as genuine or automated. The dataset provides information on the follower and friend relationships between these users.
\item MGTAB is a dataset built upon the largest raw data in the field of machine account detection, containing more than 1.5 million users and 130 million tweets. The dataset provides information on 7 types of relationships between these users and labels 10,199 accounts as either genuine or bots.
\end{itemize}

We construct user social graphs using all labeled users for the above datasets. Specifically, for MGTAB, we use the 20 user attribute features with the highest information gain and the 768-dimensional user tweet features extracted by BERT as user features. For Twibot-20, following the processing in~\cite{article16}, we use 16 user attribute features, the 768-dimensional user description features extracted by BERT, and the user tweet features. For Cresci-15, following the processing approach in~\cite{article16}, we use 6 user attribute features, the 769-dimensional user description features extracted by BERT, and the user tweet features. The statistics of these datasets are summarized in Table \ref{tb:Statistics}. We conduct a 1:1:8 random partition as training, validation, and test set for all datasets.

\begin{table}[h]
\caption{Statistics of datasets used in the paper.}
\begin{center}
\begin{adjustbox}{width=0.80\linewidth}
\begin{tabular}{lcccccc}
\toprule
Dataset & Nodes & Edges &Features &$\alpha$ &$\beta$ &$\gamma$ \\

\hline
Cresci-15  &5,301   &14,220     &1,542 &0.95 &0.95 &0.95 \\
Twibot-20  &11,826	&15,434	    &1,553 &0.8	 &0.8  &0.9 \\
MGTAB      & 10,199	&1,700,108	&788   &0.6	 &0.9  &0.8 \\
\bottomrule
\end{tabular}
\end{adjustbox}
\end{center}
\label{tb:Statistics}
\end{table}

\subsection{Baseline Methods}
To verify the effectiveness of our proposed RF-GNN, we compare it with various semi-supervised learning baselines. The detail about these baselines as described as follows:

\begin{itemize}
\item \textbf{DT}~\cite{article05} is a classification rule that infers a tree-shaped structure from a set of training data using a recursive top-down approach. It compares attribute values of the nodes inside the tree and determines the branching direction based on different attribute values.
\item \textbf{RF}~\cite{article04} uses DT as models in bagging. First, different training sets are generated using the bootstrap method. Then, for each training set, a DT is constructed, and finally, the trees are integrated into a forest and used to predict the final results.
\item \textbf{Node2Vec}~\cite{article27} is a weighted random walk algorithm that allows the trained vectors to simultaneously satisfy the homophily and structural similarity assumptions between nodes.
\item \textbf{APPNP}~\cite{article14} combines GCN with PageRank~\cite{article31} to construct a simple model that utilizes the propagation of PageRank. It uses a large, adjustable neighborhood to better propagate information from neighboring nodes.
\item \textbf{GCN}~\cite{article17} is a representative of the spectral graph convolution method. By simplifying Chebyshev polynomials to first order neighborhood, the node embedding vector is obtained.
\item \textbf{SGC}~\cite{article18} is a simplified version of the GCN. It aims to decrease the excessive complexity of GCNs by iteratively removing non-linearities between GCN layers and collapsing the resulting function into a single linear transformation. Utilizing this technique ensures that the performance is comparable to that of GCNs, while significantly reducing the parameter size.
\item \textbf{GAT}~\cite{article19} is a semi-supervised homogeneous graph model that applies the attention mechanism to determine the weights of node neighborhood. By adaptively assigning weights to different neighbors, the performance of graph neural networks is improved.
\item \textbf{Boosting-GNN}~\cite{article20} is a graph ensemble learning method that combines GNN with adaboost, and can improve the performance of GNN under class imbalance conditions.
\item \textbf{JK-Nets}~\cite{article21} is a kind of GNN that employs jump knowledge to obtain a more effective structure-aware representation by flexibly utilizing the distinct neighborhood ranges of each node.
\item \textbf{GraphSAINT}~\cite{article22} is an inductive learning approach that relies on graph sampling. It addresses the issue of neighbor explosion by sampling subgraphs and applying GCN on them, while ensuring minimal variance and unbiasedness.
\item \textbf{LA-GCN}~\cite{article23} improves the expressiveness of GNN by generating neighborhood features based on a conditional generative model, which takes into account the local structure and node features.
\end{itemize}

\subsection{Variants}
To gain a more comprehensive understanding of how each module operates within the overall learning framework and to better evaluate their individual contributions to performance improvement, we generated several variants of the full RF-GNN model. The RF-GNN model consists of three primary modules: subgraph construction module, the alignment mechanism, and model ensemble module. To conduct an ablation study, we selectively enabled or disabled certain components of these modules. The following is a detailed description of these variations:

\begin{itemize}
\item \textbf{RF-GNN-\textit{E}:} This variant solely employs emsembling to determine its effects without utilizing any other modules. The base classifier is trained using the entire training set every time. The final classification result is obtained by aggregating the outputs of the base classifiers using bagging. In this model variant, the number of base classifiers is fixed to be $S$.
\item \textbf{RF-GNN-\textit{ES}:} This variant utilizes the bagging method, together with the subgraph construction. To increase the diversity of the base classifiers, $S$ different subgraphs are obtained through subgraph construction before training the base classifiers. Each subgraph is used to train one of the $S$ base classifiers.
\item \textbf{RF-GNN} contains all modules in the graph learning framework. The aligning mechanism is additionally supplemented upon RF-GNN-\textit{ES}.
\end{itemize}

\subsection{Parameter Settings}
We train all models using AdamW optimizer for 200 epochs. The learning rate is set to 0.01 for all models with an exception to 0.005 for Node2Vec. The L2 weight decay factor is set to 5e-4 on all datasets. The dropout rate is set from 0.3 to 0.5. For all models, the input and output dimensions of the GNN layers are consistent, which are 128 or 256. Attention heads for GAT and RGAT~\cite{article30} are set to 4. We implement RF-GNN with Pytorch 1.8.0, Python 3.7.10, PyTorch Geometric~\cite{article34} with sparse matrix multiplication. All experiments are executed on a sever with 9 Titan RTX GPU, 2.20GHz Intel Xeon Silver 4210 CPU with 512GB RAM. The operating system is Linux bcm 3.10.0.

\subsection{Evaluation Metrics}
Since the number of humans and bots is not roughly equal on social media, we utilize Accuracy and F1-score to indicate the overall performance of classifier:

\begin{equation}
\label{eq:Accuracy}
\text{Acurracy}=\frac{TP+TN}{TP+FP+FN+TN},
\end{equation}

\begin{equation}
\label{eq:Precision}
\text{Precision}=\frac{TP}{TP+FP},
\end{equation}

\begin{equation}
\label{eq:Recall}
\text{Recall}=\frac{TP}{TP+FN},
\end{equation}

\begin{equation}
\label{eq:F1}
\text{F1}=\frac{2 \times \text{Precision} \times \text {Recall}}{\text {Precision}+ \text {Recall}},
\end{equation}

where $TP$ is True Positive, $TN$ is True Negative, $FP$ is False Positive, $FN$ is False Negative.

\section{Experiment results}
In this section, we conduct several experiments to evaluate RF-GNN. We mainly answers the following questions:
\begin{itemize}
\item \textbf{Q1:} How different algorithms perform in different scenarios, i.e., algorithm effectiveness (Section \ref{sec:Overall_performance}).
\item \textbf{Q2:} How each individual module of RF-GNN contributes to the overall effectiveness (Section \ref{sec:Ablation}).
\item \textbf{Q3:} How does the number of base classifiers effect the performance of RF-GNN. (Section \ref{sec:base_classifier}).
\item \textbf{Q4:} How is RF-GNN perform under different parameters setting. i.e., parameter sensitivity. (Section \ref{sec:Sensitivity}).
\item \textbf{Q5:} How does RF-GNN perform when applied to heterogeneous GNNs. i.e., extensibility (Section \ref{sec:extensibility}).
\item \textbf{Q6:} How different algorithms perform when dealing data containing noise, i.e., robustness (Section \ref{sec:Robustness}).
\end{itemize}

\subsection{Overall performance}
\label{sec:Overall_performance}
In this section, we perform experiments on public available social bot detection datasets to evaluate the effectiveness of our proposed method. We randomly conduct a 1:1:8 partition as training, validation, and test set. To reduce randomness and ensure the stability of the results, each method was evaluated five times with different seeds. We report the average test results of baselines, RF-GNN, and the variants. As shown in Table \ref{tb:main_results}, RF-GNN outperforms other baselines and different variants across all scenarios.

\begin{table*}[t]
\caption{Comparison of the average performance of different methods for social bot detection. For RF-GNN, $S$ is set to 10 for all dataset. The best result of the baseline method and the complete RF-GNN method proposed by us is highlighted in bold.}
\begin{center}
\begin{scriptsize}
\setlength{\columnsep}{1pt}%
\begin{adjustbox}{width=0.965\linewidth}
\begin{tabular}{@{\extracolsep{1pt}}rlcc|cc|cc@{}}
\toprule
 & \multirow{2}{*}{\textbf{Method}} & \multicolumn{2}{c}{MGTAB} & \multicolumn{2}{c}{Twibot-20} & \multicolumn{2}{c}{Cresci-15}  \\
\cline{3-8}
\rule{0pt}{2.2ex}
& & Acc & F1 & Acc & F1 & Acc & F1 \\
\cline{2-8}
\rule{0pt}{2.5ex}
\multirow{11}{*}{\rotatebox{90}{Baseline}}
& Decision Tree    & 78.80 \tiny{$\pm 1.46$} & 73.22 \tiny{$\pm 1.89$} & 65.00 \tiny{$\pm 6.27$}
                   & 64.67 \tiny{$\pm 6.32$} & 88.42 \tiny{$\pm 1.61$} & 87.01 \tiny{$\pm 2.01$}
             \\
& Random Forest    & 84.46 \tiny{$\pm 1.33$} & 78.45 \tiny{$\pm 1.95$} & 73.42 \tiny{$\pm 1.21$}
                   & 73.29 \tiny{$\pm 1.25$} & 94.36 \tiny{$\pm 0.52$} & 93.92 \tiny{$\pm 0.53$}
             \\
\cdashline{2-8}
\rule{0pt}{2.5ex}
& Node2Vec         & 73.35 \tiny{$\pm 0.19$} & 60.20 \tiny{$\pm 0.72$} & 51.85 \tiny{$\pm 0.20$}
                   & 48.98 \tiny{$\pm 0.39$} & 73.22 \tiny{$\pm 0.60$} & 70.83 \tiny{$\pm 0.56$}
             \\
& APPNP            & 75.08 \tiny{$\pm 1.73$} & 61.66 \tiny{$\pm 1.25$} & 53.13 \tiny{$\pm 3.80$}
                   & 50.82 \tiny{$\pm 3.38$} & 95.33 \tiny{$\pm 0.48$} & 94.97 \tiny{$\pm 0.51$}
             \\
& GCN              & 84.98 \tiny{$\pm 0.70$} & 79.63 \tiny{$\pm 1.02$} & 67.76 \tiny{$\pm 1.24$}
                   & 67.34 \tiny{$\pm 1.16$} & 95.19 \tiny{$\pm 0.99$} & 94.88 \tiny{$\pm 1.02$}
             \\
& SGC              & 85.14 \tiny{$\pm 0.72$} & 80.60 \tiny{$\pm 1.66$} & 68.01 \tiny{$\pm 0.40$}
                   & 67.60 \tiny{$\pm 0.24$} & 95.69 \tiny{$\pm 0.84$} & 95.39 \tiny{$\pm 0.85$}
             \\
& GAT              & 84.94 \tiny{$\pm 0.29$} & 80.22 \tiny{$\pm 0.44$} & 71.71 \tiny{$\pm 1.36$}
                   & 71.18 \tiny{$\pm 1.38$} & 96.10 \tiny{$\pm 0.46$} & 95.79 \tiny{$\pm 0.49$}
             \\
& Boosting-GNN     & 85.14 \tiny{$\pm 0.72$} & 79.84 \tiny{$\pm 1.09$} & 68.10 \tiny{$\pm 0.77$}
                   & 67.77 \tiny{$\pm 0.79$} & 95.69 \tiny{$\pm 0.47$} & 95.40 \tiny{$\pm 0.49$}
             \\
& JK-Nets          & 84.58 \tiny{$\pm 0.28$} & 80.60 \tiny{$\pm 0.78$} & 71.01 \tiny{$\pm 0.54$}
                   & 70.77 \tiny{$\pm 0.37$} & 96.04 \tiny{$\pm 0.42$} & 95.76 \tiny{$\pm 0.43$}
             \\
& GraphSAINT       & 84.38 \tiny{$\pm 0.11$} & 79.06 \tiny{$\pm 0.19$} & \textbf{76.49 \tiny{$\pm 0.06$}}
                   & \textbf{76.04 \tiny{$\pm 0.14$}} & \textbf{96.18 \tiny{$\pm 0.07$}} & \textbf{95.91 \tiny{$\pm 0.07$}}
             \\
& LA-GCN           & \textbf{85.50 \tiny{$\pm 0.28$}} & \textbf{81.12 \tiny{$\pm 0.42$}} & 74.36 \tiny{$\pm 0.67$}
                   & 73.49 \tiny{$\pm 0.67$} & 96.02 \tiny{$\pm 0.39$} & 95.70 \tiny{$\pm 0.43$}
             \\
\cline{2-8}
\noalign{\vskip\doublerulesep
         \vskip-\arrayrulewidth} \cline{2-8}
\rule{0pt}{2.5ex}
\multirow{9}{*}{\rotatebox{90}{Ablation}}
& RF-GCN-\textit{E} & 84.75 \tiny{$\pm 1.02$} & 79.47 \tiny{$\pm 1.89$} & 72.81 \tiny{$\pm 1.11$}
                    & 72.17 \tiny{$\pm 1.07$} & 95.23 \tiny{$\pm 1.07$} & 94.92 \tiny{$\pm 1.10$}
             \\
& RF-GCN-\textit{ES}& 84.73 \tiny{$\pm 0.12$} & 79.30 \tiny{$\pm 0.71$} & 73.12 \tiny{$\pm 0.67$}
                    & 72.74 \tiny{$\pm 0.74$} & 96.02 \tiny{$\pm 0.31$} & 95.72 \tiny{$\pm 0.32$}
             \\
& RF-GCN            & \textbf{86.99 \tiny{$\pm 0.20$}} & \textbf{82.94 \tiny{$\pm 0.45$}} & \textbf{82.21 \tiny{$\pm 0.51$}}
                    & \textbf{81.85 \tiny{$\pm 0.53$}} & \textbf{96.40 \tiny{$\pm 0.12$}} & \textbf{96.10 \tiny{$\pm 0.13$}}
             \\
\cdashline{2-8}
\rule{0pt}{2.5ex}
& RF-SGC-\textit{E} & 85.23 \tiny{$\pm 0.64$} & 80.23 \tiny{$\pm 1.32$} & 70.76 \tiny{$\pm 0.63$}
                    & 70.12 \tiny{$\pm 0.80$} & 95.94 \tiny{$\pm 0.72$} & 95.64 \tiny{$\pm 0.74$}
             \\
& RF-SGC-\textit{ES}& 85.68 \tiny{$\pm 0.19$} & 80.82 \tiny{$\pm 0.34$} & 70.36 \tiny{$\pm 0.56$}
                    & 69.87 \tiny{$\pm 0.70$} & 96.26 \tiny{$\pm 0.18$} & 95.96 \tiny{$\pm 0.20$}
             \\
& RF-SGC            & \textbf{86.96 \tiny{$\pm 0.58$}} & \textbf{83.36 \tiny{$\pm 0.58$}} & \textbf{81.98 \tiny{$\pm 0.45$}}
                    & \textbf{81.65 \tiny{$\pm 0.46$}} & \textbf{96.43 \tiny{$\pm 0.16$}} & \textbf{96.13 \tiny{$\pm 0.18$}}
             \\
\cdashline{2-8}
\rule{0pt}{2.5ex}
& RF-GAT-\textit{E} & 85.76 \tiny{$\pm 0.15$} & 81.07 \tiny{$\pm 0.54$} & 72.63 \tiny{$\pm 1.06$}
                    & 72.11 \tiny{$\pm 1.05$} & 96.56 \tiny{$\pm 0.05$} & 96.28 \tiny{$\pm 0.06$}
             \\
& RF-GAT-\textit{ES}       & 84.90 \tiny{$\pm 0.03$} & 79.30 \tiny{$\pm 0.07$} & 74.07 \tiny{$\pm 0.60$}
                    & 73.76 \tiny{$\pm 0.64$} & 96.45 \tiny{$\pm 0.05$} & 96.16 \tiny{$\pm 0.05$}
             \\
& RF-GAT           & \textbf{86.56 \tiny{$\pm 0.12$}} & \textbf{83.14 \tiny{$\pm 0.23$}} & \textbf{81.69 \tiny{$\pm 0.32$}}
                   & \textbf{81.34 \tiny{$\pm 0.37$}} & \textbf{96.62 \tiny{$\pm 0.05$}} & \textbf{96.33 \tiny{$\pm 0.05$}}
             \\
\bottomrule
\end{tabular}
\end{adjustbox}
\end{scriptsize}
\end{center}
\label{tb:main_results}
\end{table*}

The performance of the RF is significantly improved compared to the base classifier DT on all datasets. However, due to the inability of RF to utilize the relationship features among users, it performs less effectively than the GCN classifier on the MGTAB and Cresci-15 datasets.

Node2Vec consistently performs poorly compared to other baseline methods across all datasets. This can be attributed to Node2Vec's approach of controlling the random walk process within the step size. It struggles to capture the similarity of adjacent nodes in large-scale graphs with highly complex structures and does not effectively utilize node features. The APPNP model incorporates PageRank computation, injecting some global information into the learning process and expanding the neighborhood size. This leads to an improvement in performance compared to Node2Vec. The GCN approach involves multiplying the normalized adjacency matrix with the feature matrix and then multiplying it with a trainable parameter matrix to perform convolution operations on the entire graph data. However, relying on full-graph convolution operations to obtain the global representation vector for a node can significantly impair the model's generalization performance. The SGC model removes the non-linear activation function from GCN. Despite a slightly reduced accuracy across all datasets, SGC can achieve similar performance compared to GCN. In contrast, GAT introduces an attention mechanism to GCN, allowing for adaptive model parameter adjustment during the convolution and feature fusion process by assigning a learnable coefficient to each edge. This improves the model's performance by making it more adaptive to the specific task.

\begin{figure}[ht]
  \centering
  \subfloat[GCN on MGTAB]{
   \begin{minipage}[b]{0.45\textwidth}
    \includegraphics[width=1.10\textwidth]{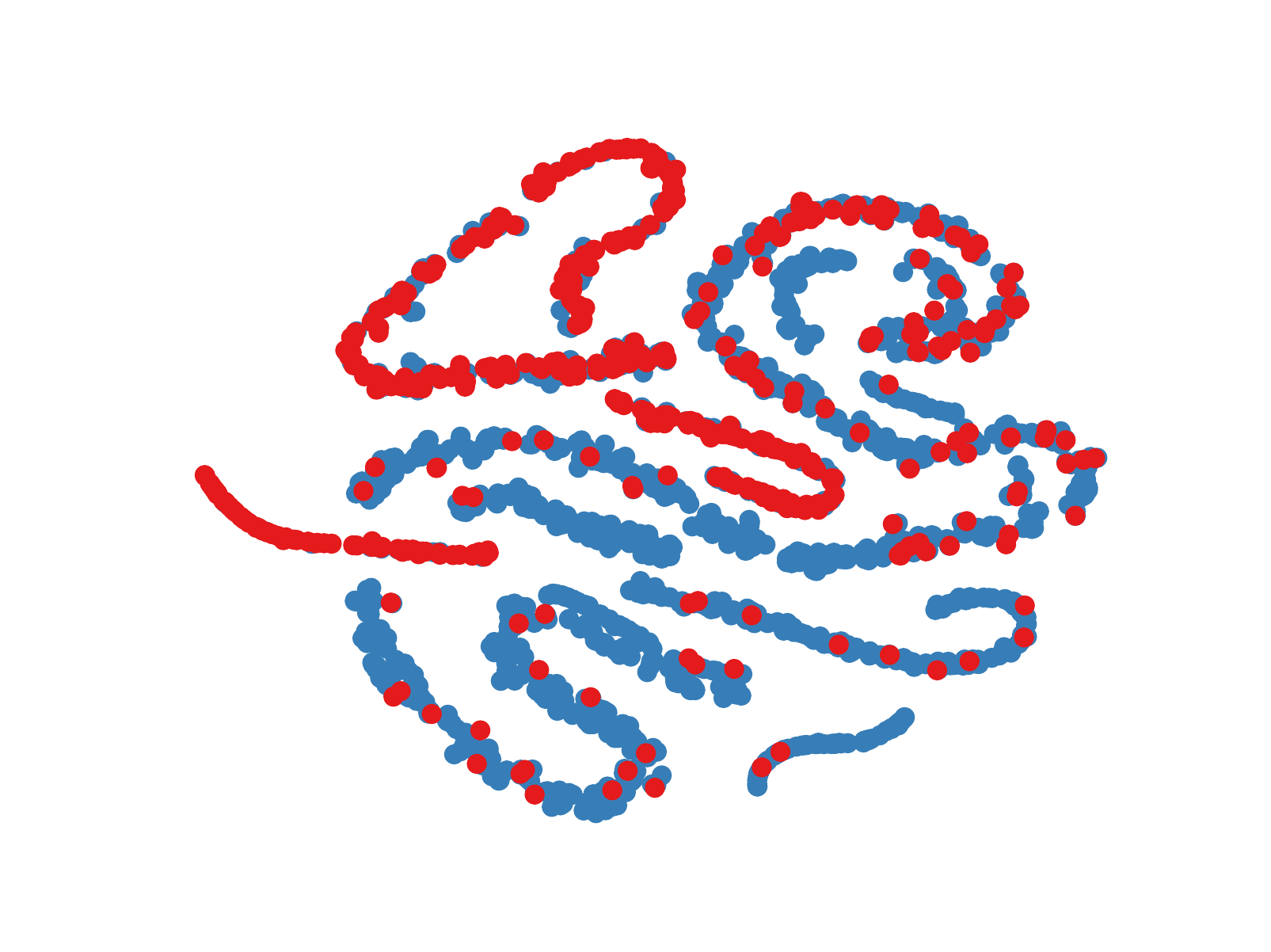}
  \end{minipage}}
  \subfloat[RF-GCN on MGTAB]{
   \begin{minipage}[b]{0.45\textwidth}
    \includegraphics[width=1.10\textwidth]{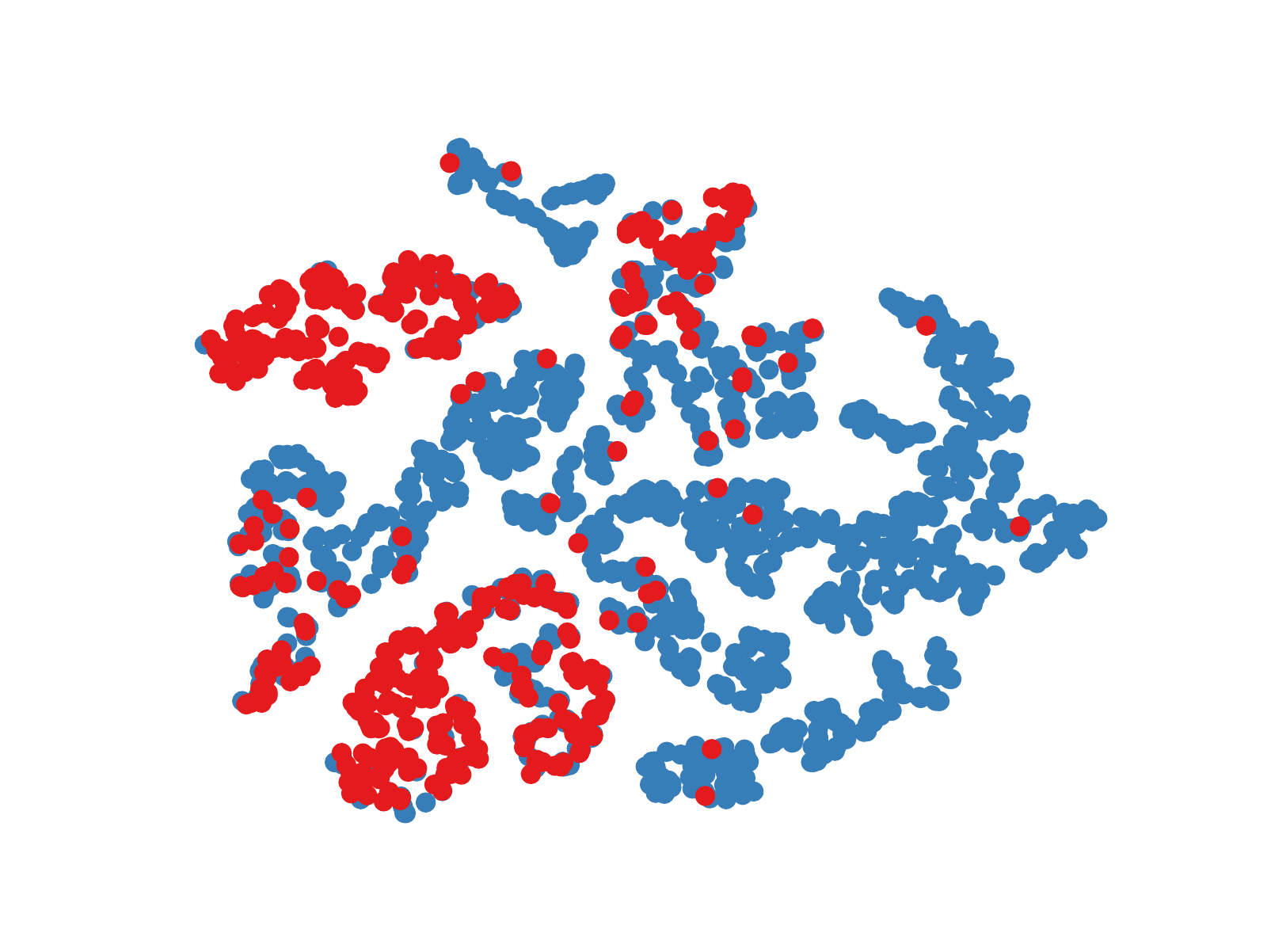}
  \end{minipage}}

  \subfloat[GCN on Twibot-20]{
   \begin{minipage}[b]{0.45\textwidth}
    \includegraphics[width=1.10\textwidth]{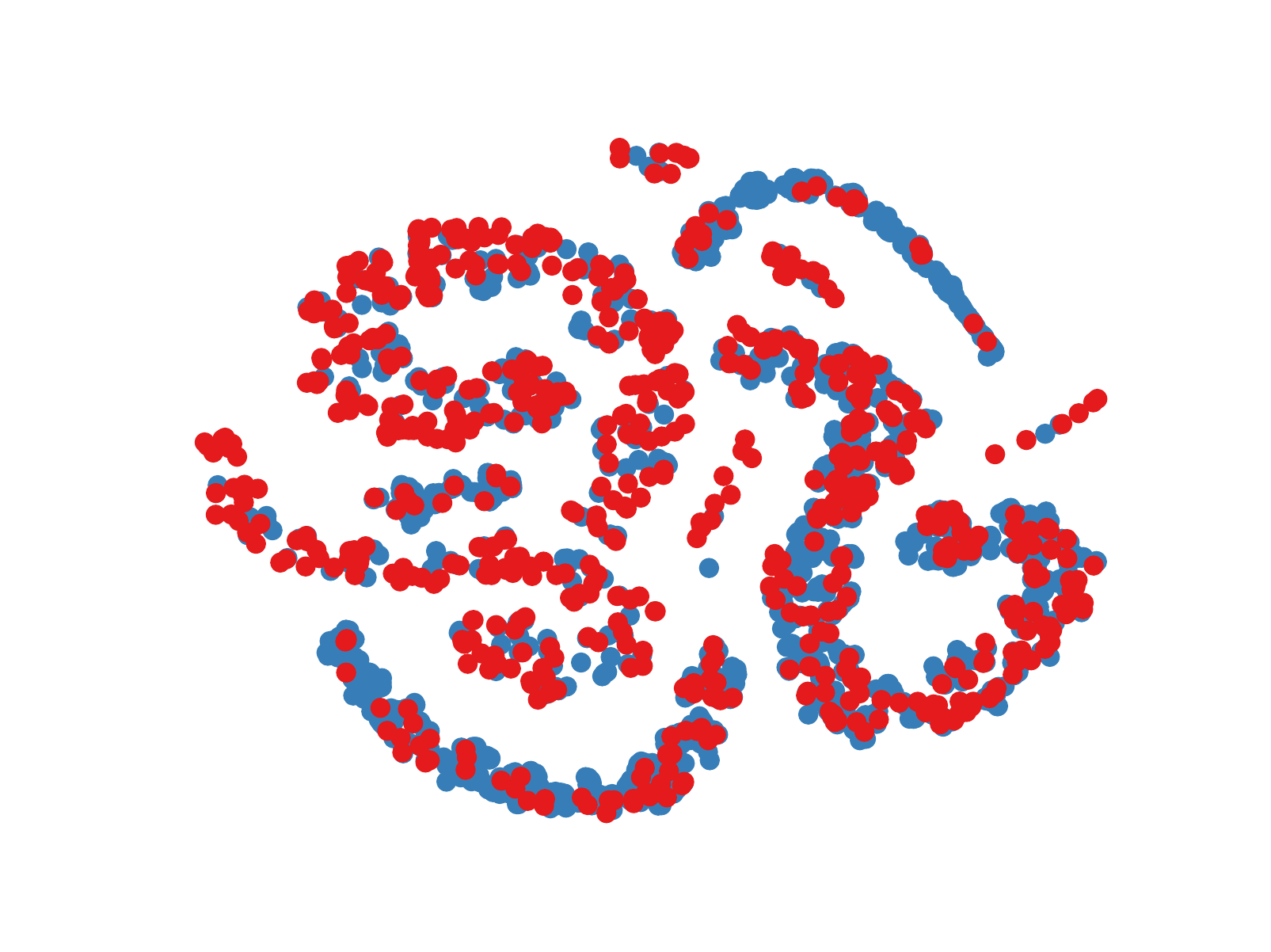}
  \end{minipage}}
  \subfloat[RF-GCN on Twibot-20]{
     \begin{minipage}[b]{0.45\textwidth}
    \includegraphics[width=1.10\textwidth]{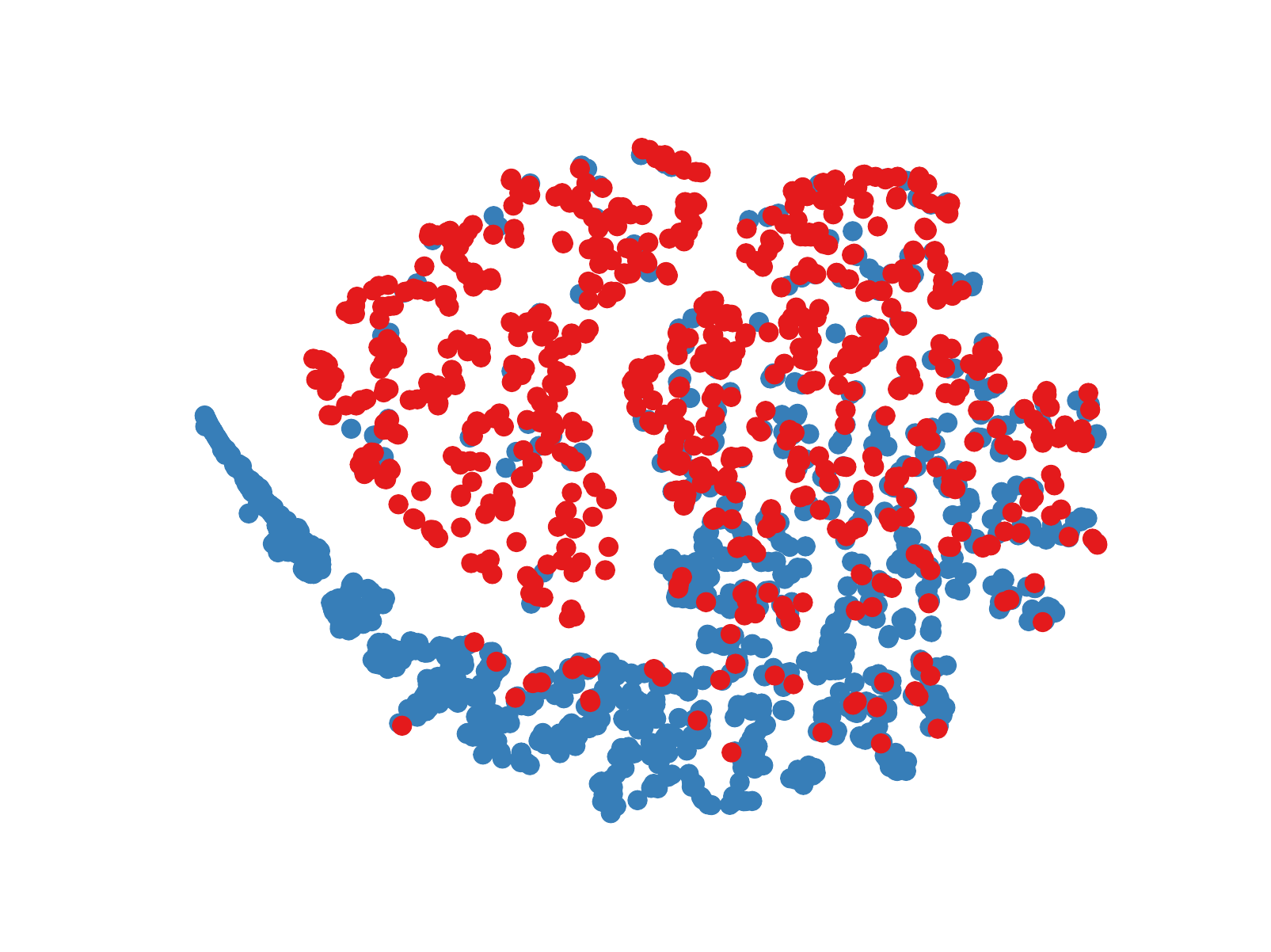}
  \end{minipage}}
  \caption{2D t-SNE visualization. 2000 nodes in dataset were randomly selected for plotting. Red and blue represent bot and human, respectively.}
  \label{fig:visualization}
\end{figure}

Bot detection on the Cresci-15 dataset is a relatively simple task, with most detection methods achieving an accuracy of over 95\%. As a result, there is little room for improvement on this dataset. Compared with the best results among all the state-of-the-arts, our method can achieve 1.38\%, 5.72\%, 0.44\% accuracy improvement, on the datasets of MGTAB, Twibot-20, and Cresci-15. Especially, for accuracy, RF-GCN, RF-SGC, RF-GAT achieves maximum improvements of 14.15\%, 13.97\%, 9.98\% on Twibot-20 compared to the baseline model GCN. The results demonstrate the effectiveness of RF-GNN. In Figure \ref{fig:visualization}, the classification capabilities of the RF-GCN model compared with GCN model on all MGTAB and Twibot-20 have been shown using t-SNE~\cite{article36} visualization.

\begin{figure}[h]
  \centering
   \begin{minipage}{0.965\textwidth}
    \centerline{\includegraphics[width=\textwidth]{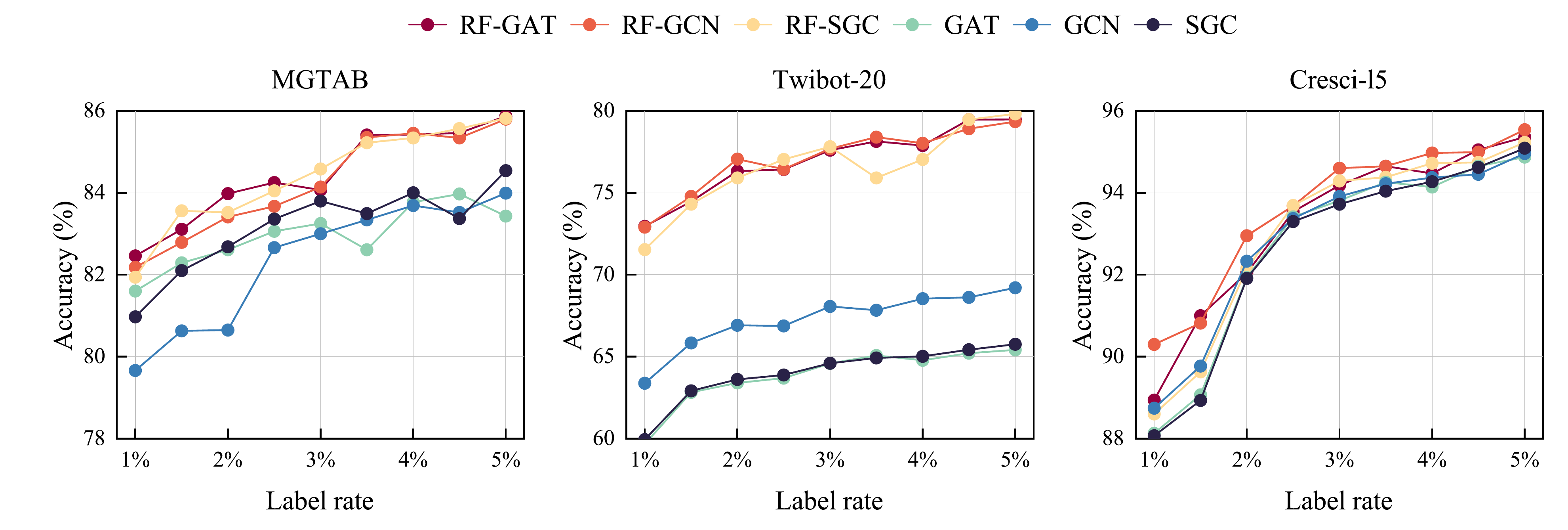}}
   \end{minipage}
  \caption{RF-GNN performance with different scale of training data.}
  \label{fig:label_rate}
\end{figure}

\noindent \textbf{Impart of scale of training data}
To further compare the performance improvement of our method, we conducted an extensive comparison between RF-GNN and the original GNN approach under various training set. Specifically, we varied the proportion of the training set from 1\% to 5\% on the social bot detection dataset. We used GCN, SGC, and GAT as the backbone models, and the results are shown in Figure \ref{fig:label_rate}.

The proposed RF-GNN outperforms the backbone models by a significant margin on all social bot detection datasets under different training dataset. On the Cresci-15, Twibot-20, and MGTAB datasets, RF-GNN achieves an average accuracy improvement of 3.38\%, 9.55\%, and 5.35\%, respectively.

\subsection{Ablation Study}
\label{sec:Ablation}
The second half of Table \ref{tb:main_results} presents the performance of various variants, highlighting the roles of different modules in our proposed learning framework. RF-GNN-\textit{E}, which only employs multi-model ensembling, exhibits a minor improvement over the original GNN model. This is because the different base classifiers use the same training set, and the differences between the different models are relatively small, which limits the potential improvement from ensembling. RF-GNN-\textit{ES}, on the other hand, improves model performance in most cases by incorporating subgraph construction and training the base classifiers on different subgraphs. However, since the constructed subgraphs only use a subset of nodes and features, information from unsampled nodes and features may be missing, leading to performance degradation in certain cases. For instance, on the MGTAB dataset, RF-GCN-\textit{ES} underperforms RF-GCN-E. As illustrated in the results of RF-GNN, combining the aligning mechanism with the backbone RF-GNN-\textit{ES} can further enhance the performance. Most noticeably, the accuracy can be substantially augmented over 13\% for GCN and SGC when tackling Twibot-20 dataset.

\begin{figure}[ht]
  \centering
   \begin{minipage}[b]{0.32\textwidth}
    \includegraphics[width=1\textwidth]{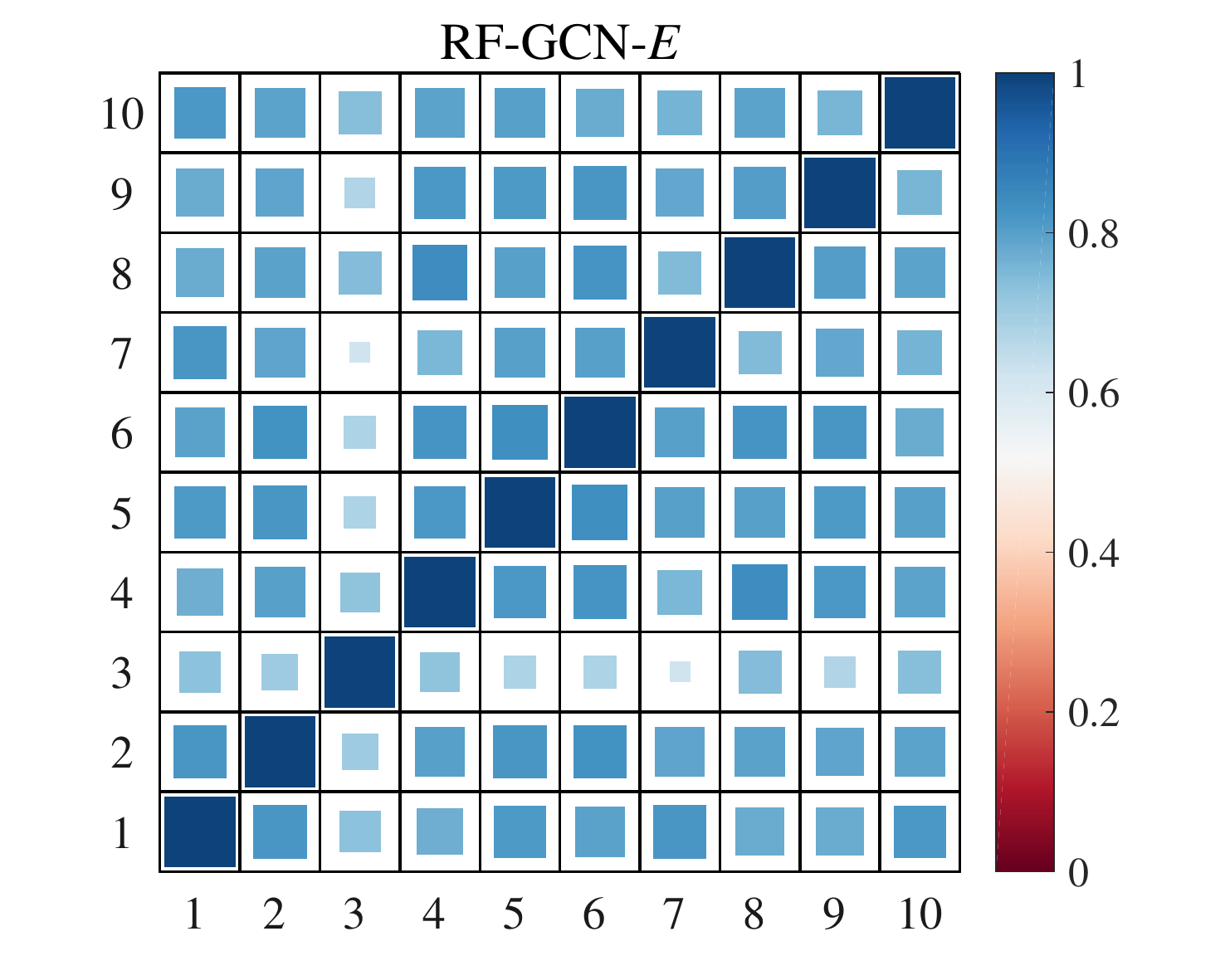}
  \end{minipage}
   \begin{minipage}[b]{0.32\textwidth}
    \includegraphics[width=1\textwidth]{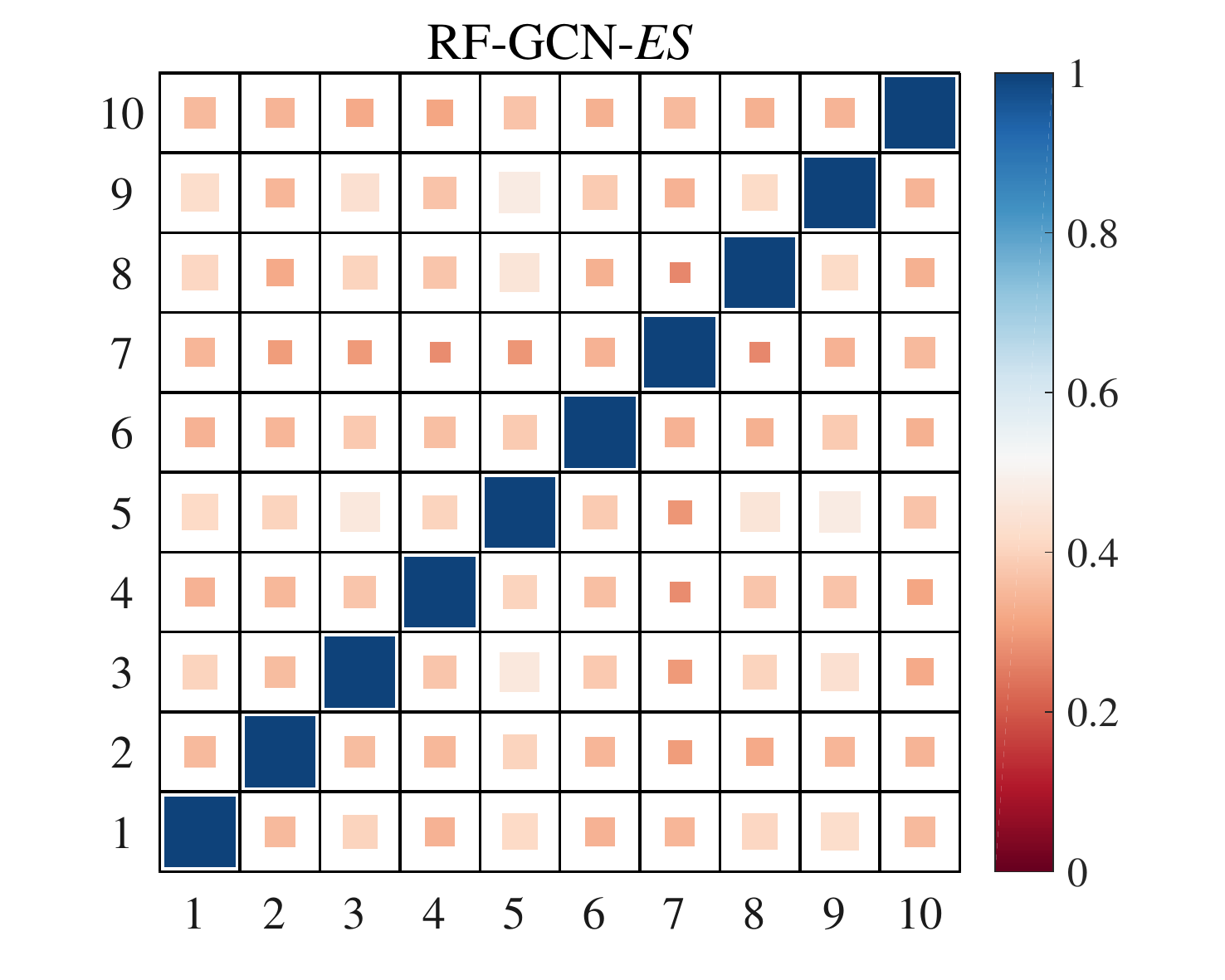}
  \end{minipage}
   \begin{minipage}[b]{0.32\textwidth}
    \includegraphics[width=1\textwidth]{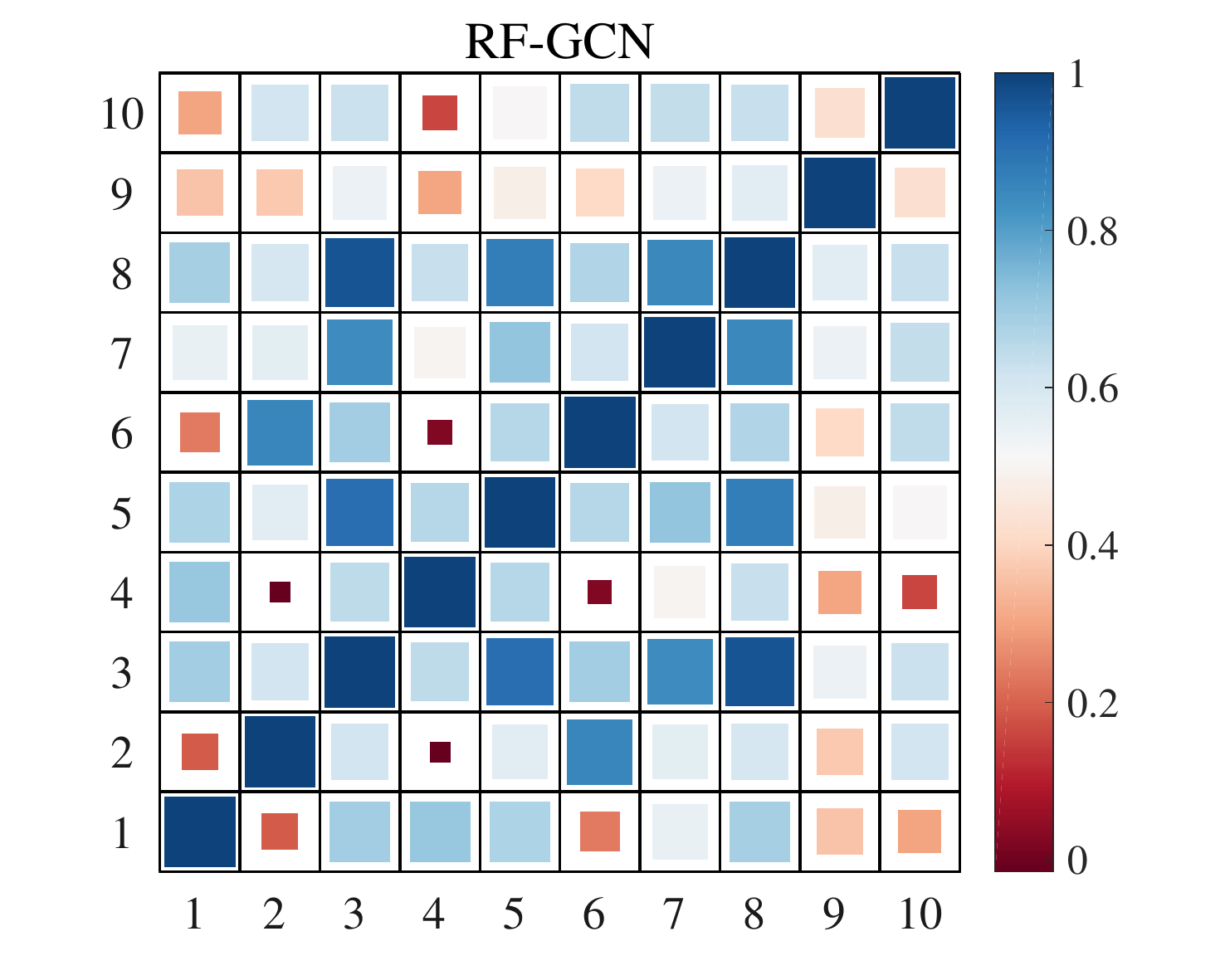}
  \end{minipage}
  \caption{Average cosine similarity of the output of the base classifiers on Twibot-20.}
  \label{fig:heat}
\end{figure}

We separately calculated the average cosine similarity of the output of the base classifiers of RF-GCN-\textit{E}, RF-GCN-\textit{ES}, and RF-GCN on Twibot-20, as shown in Figure \ref{fig:heat}. Higher similarity in the output indicates higher consistency among the base classifiers, resulting in more accurate results for the ensemble model. Although the output similarity of the base classifiers of RF-GCN-\textit{E} is high, the base classifiers of RF-GCN-\textit{E} are trained on the same training set, resulting in limited diversity and a small gain in integrating the output results of the base classifiers. For the RF-GCN-\textit{ES}, the subgraph construction ensures the differences between the base classifiers, but the low output similarity suggests that the accuracy of each base classifier is not high. Using the same subgraph training as RF-GCN-\textit{ES}, RF-GCN significantly improves the output similarity of different branches by introducing the alignment mechanism.

\subsection{The effect of the number of base classifiers}
\label{sec:base_classifier}
The number of base classifiers $S$ is a crucial parameter that impacts the performance of a RF-GNN. With a small number of base classifiers, the classification error of the RF-GNN is high, resulting in relatively poor performance.
Although increasing $S$ can ensure the diversity of the ensemble classifier, the construction time of the RF-GNN is directly proportional to $S$, setting $S$ too large can lead to a decrease in model efficiency.

\begin{figure}[h]
  \centering
   \begin{minipage}{0.965\textwidth}
    \centerline{\includegraphics[width=\textwidth]{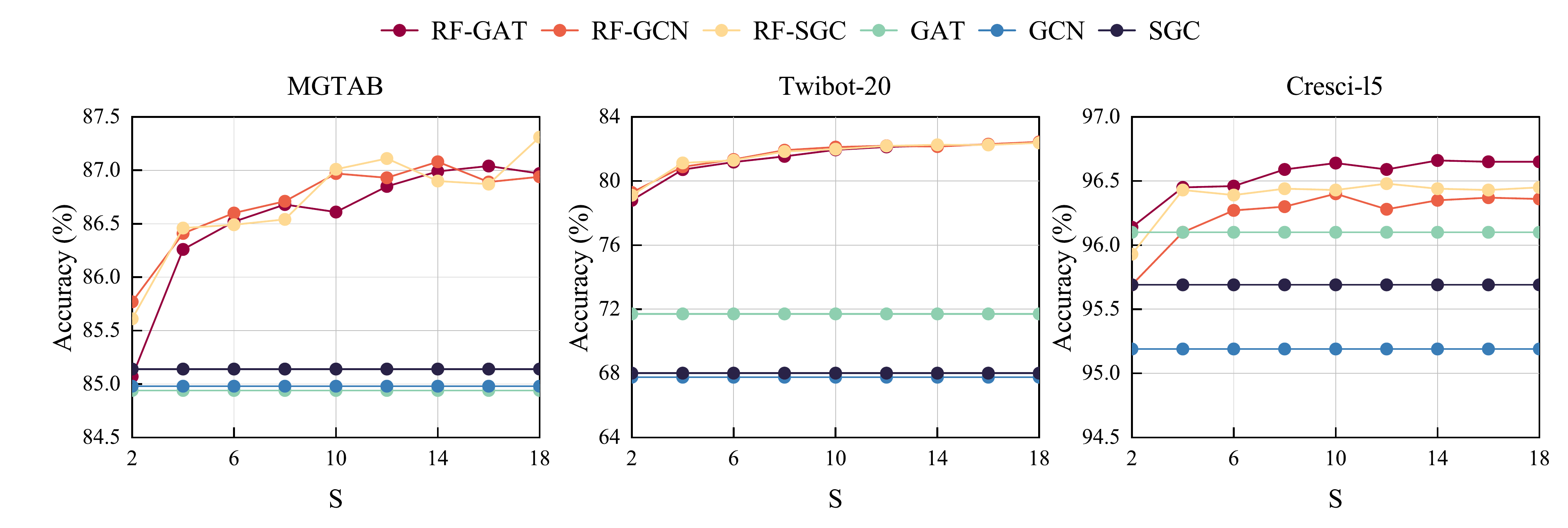}}
   \end{minipage}
  \caption{RF-GNN performance changing trend w.r.t $N$.}
  \label{fig:base_classifier}
\end{figure}

As shown in Figure \ref{fig:base_classifier}, for all datasets without exception, all the model instances experience a rise of accuracy when the number of base classifiers grows. On the MGTAB dataset, the accuracy significantly increases as $S$ increases from 2 to 10. On the Twibot-20 and Cresci-15 datasets, the accuracy shows an increasing trend as $S$ is less than 6. Across all datasets, the trend of increasing classification accuracy of RF-GNN slows down when the $S$ increases to 10.

\subsection{Parameters Sensitivity Analysis}
\label{sec:Sensitivity}
\begin{figure}[ht]
  \centering
   \begin{minipage}[b]{0.31\textwidth}
    \includegraphics[width=1.10\textwidth]{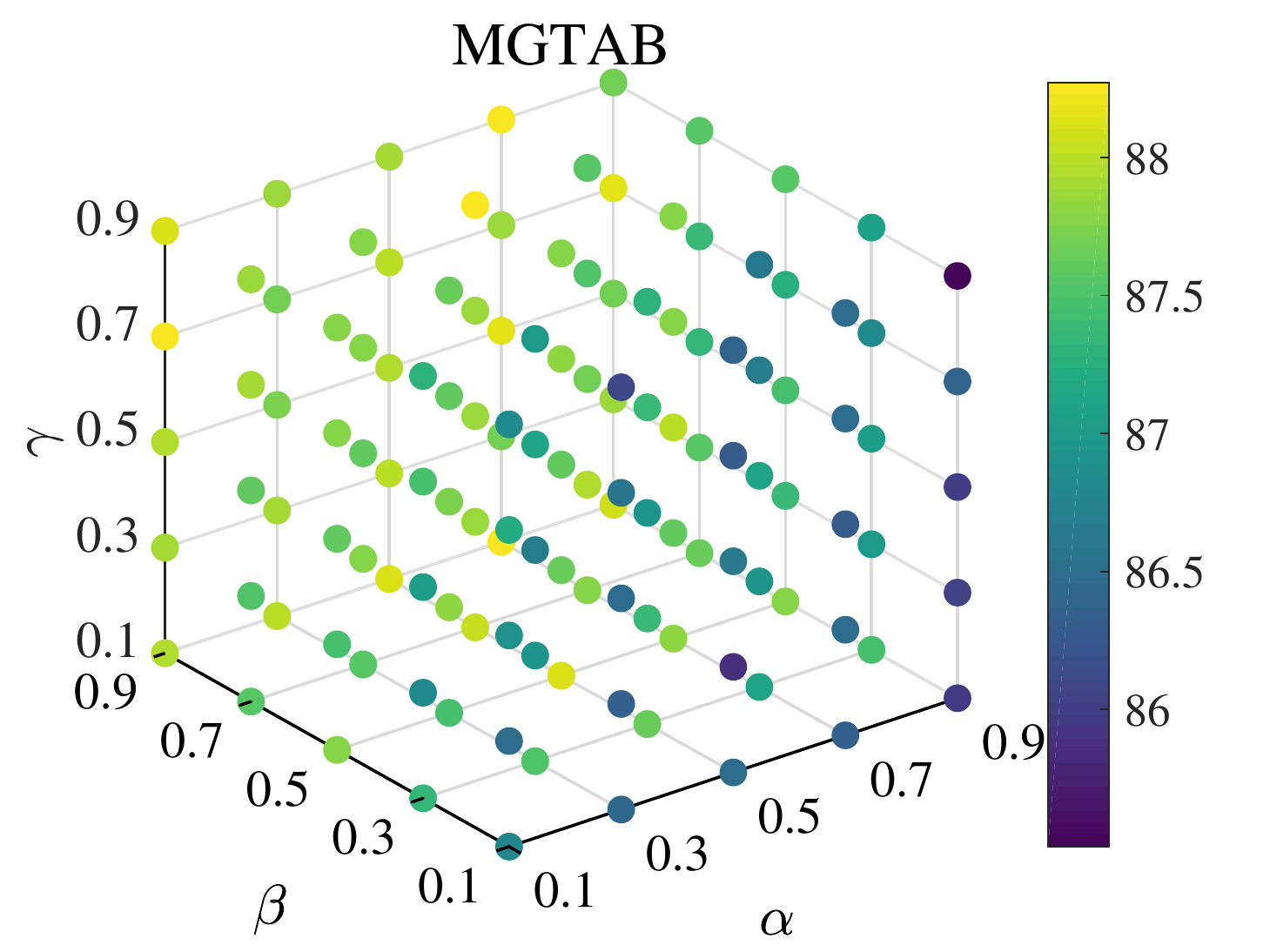}
  \end{minipage}
   \begin{minipage}[b]{0.31\textwidth}
    \includegraphics[width=1.10\textwidth]{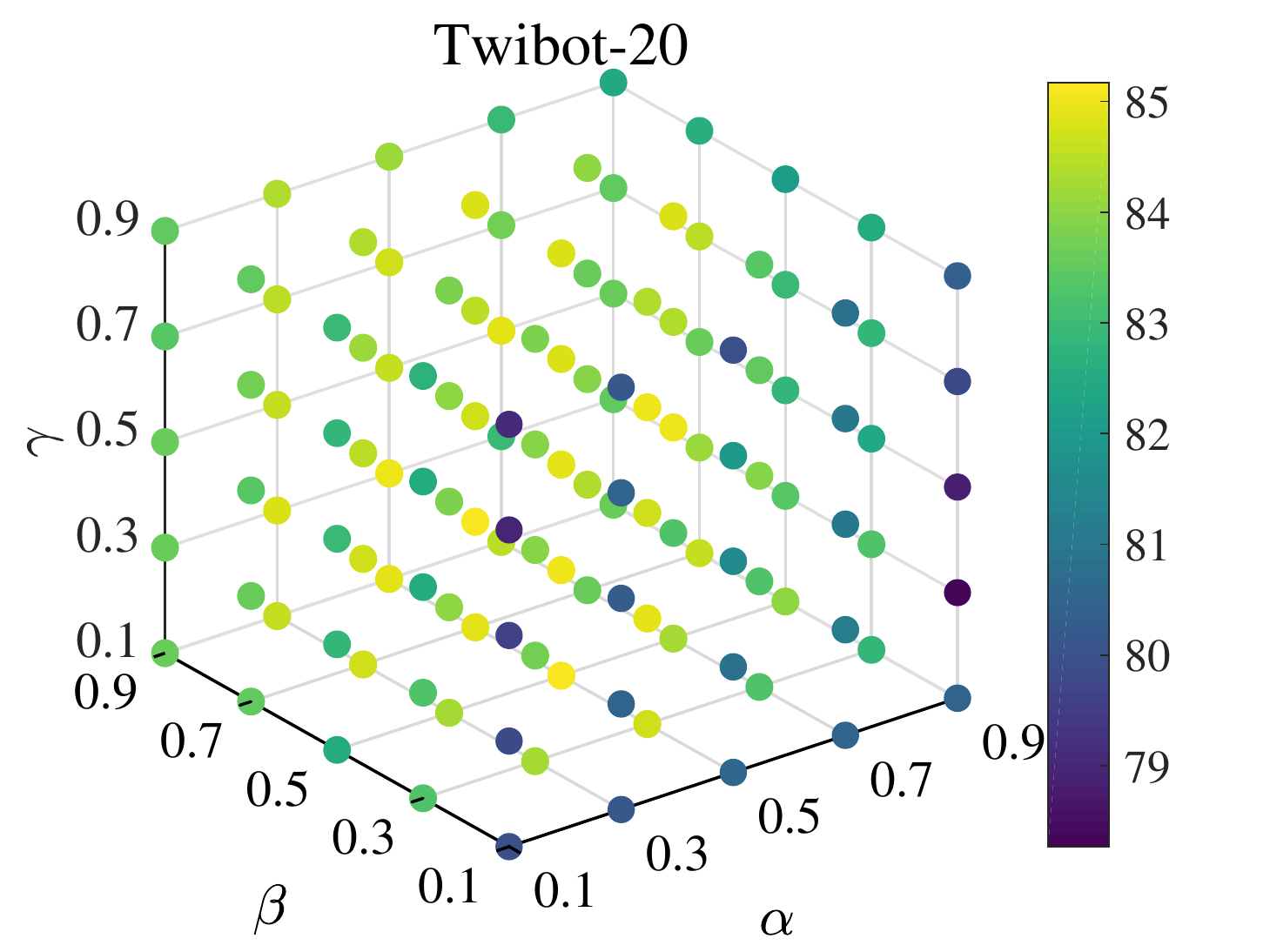}
  \end{minipage}
   \begin{minipage}[b]{0.31\textwidth}
    \includegraphics[width=1.10\textwidth]{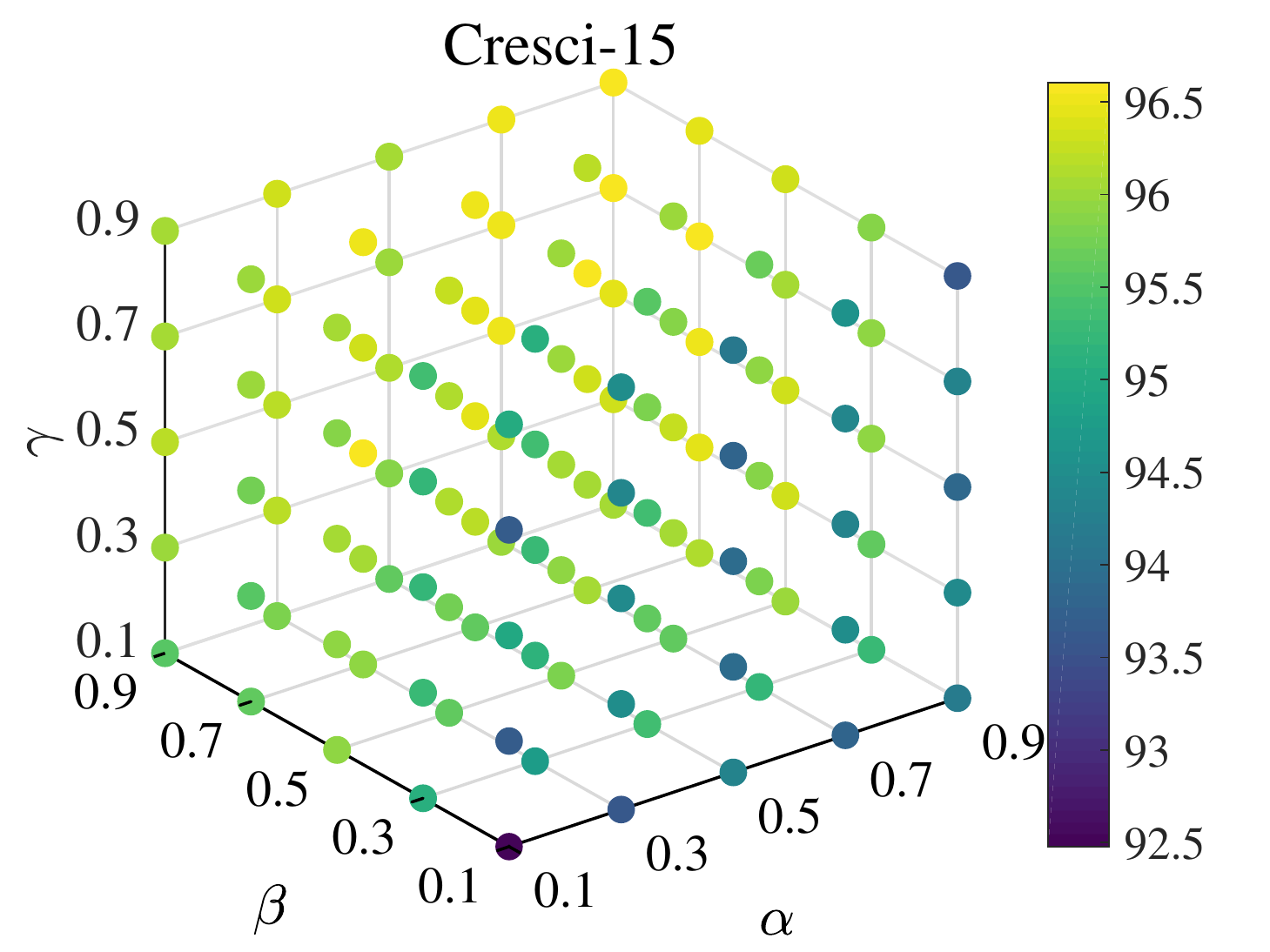}
  \end{minipage}
  \caption{The performance of RF-GCN with varying different hyperparameters in terms of accuracy.}
  \label{fig:parameter_analysis}
\end{figure}

We investigate the hyper-parameters sensitivity based on the GCN backbone models. The hyperparameters include $\alpha$, which adjusts the proportion of nodes selected when constructing subgraphs, $\beta$, which adjusts the proportion of features selected, and $\gamma$, which adjusts the proportion of edges retained between nodes. We test the above hyperparameters, and vary them from 0.1 to 0.9, the results are shown in Figure \ref{fig:parameter_analysis}.

\noindent \textbf{Analysis of node sampling probability $\alpha$}
On both MGTAB and Twibot-20 datasets, increasing the node sampling probability $\alpha$ leads to an initial improvement in performance which is followed by a gradual decline. The optimal value of a for RF-GCN is 0.5, yielding the best performance. On the Cresci-15 dataset, the performance of RF-GCN improves gradually with the increase of $\alpha$. Overall, RF-GCN exhibits stability when $\alpha$ is within the range of 0.3 to 0.7 across all datasets.

\noindent \textbf{Analysis of feature selecting probability $\beta$}
The feature selection ratio is the parameter that has the greatest impact on the performance of RF-GCN. Across all datasets, when $\beta$ is less than 0.3, the performance of RF-GCN is poor. This is likely due to the insufficient number of features, which hinders the model's ability to effectively learn and detect bots. On the Twibot-20 dataset, RF-GCN performs best when $\beta$ is set to 0.7. On the MGTAB and Cresci-15 datasets, the optimal value of $\beta$ is 0.9 for achieving the best performance.

\noindent \textbf{Analysis of edge keeping probability $\gamma$}
RF-GCN is stable when the $\gamma$ is within the range from 0.1 to 0.9 on all datasets. Reducing the value of $\gamma$ will increase the effect of data enhancement on the graph data, and the model performance will be slightly improved.

\subsection{Extend to heterogeneous GNNs}
\label{sec:extensibility}
The Cresci-15, Twibot-20, and MGTAB datasets all contain at least two types of relationships, including followers and friends. Although both types of relationships were used in Section \ref{sec:Overall_performance}, homogeneous graph models employed did not distinguish between different types of relationships, edges in graph, during neighborhood aggregation. Homogeneous GNNs can effectively utilize multiple relationships to achieve better detection results. In this section, we extend the RF-GNN framework to heterogeneous GNNs, specifically using RGCN~\cite{article29} and RGAT~\cite{article30} as base classifiers, with results shown in Table \ref{tb:extensibility}.

\begin{table*}[t]
\caption{Comparison of the average performance of homogeneous GNNs for social bot detection. Each method was evaluated five times with different seeds. The best result of the baseline method and the complete RF-GNN method proposed by us is highlighted in bold.}
\begin{center}
\begin{scriptsize}
\setlength{\columnsep}{1pt}%
\begin{adjustbox}{width=0.965\linewidth}
\begin{tabular}{@{\extracolsep{1pt}}lcc|cc|cc@{}}
\toprule
\multirow{2}{*}{\textbf{\text{ }Method}} & \multicolumn{2}{c}{MGTAB} & \multicolumn{2}{c}{Twibot-20} & \multicolumn{2}{c}{Cresci-15}  \\
\cline{2-7}
\rule{0pt}{2.2ex}
& Acc & F1 & Acc & F1 & Acc & F1 \\
\cline{1-7}
\rule{0pt}{2.5ex}
RGCN               & 86.50 \tiny{$\pm 0.49$} & 82.46 \tiny{$\pm 0.68$} & 80.10 \tiny{$\pm 0.49$}
                   & 79.76 \tiny{$\pm 0.46$} & 96.42 \tiny{$\pm 0.14$} & 96.14 \tiny{$\pm 0.15$}
             \\
\cdashline{1-7}
\rule{0pt}{2.5ex}
RF-RGCN-\textit{E} & 86.69 \tiny{$\pm 0.36$} & 82.74 \tiny{$\pm 0.49$} & 80.29 \tiny{$\pm 0.58$}
                   & 79.96 \tiny{$\pm 0.54$} & 96.46 \tiny{$\pm 0.31$} & 96.17 \tiny{$\pm 0.33$}
             \\
\text{ }RF-RGCN-\textit{ES}& 87.25 \tiny{$\pm 0.39$} & 83.20 \tiny{$\pm 0.54$} & 82.32 \tiny{$\pm 0.48$}
                   & 82.01 \tiny{$\pm 0.47$} & 96.51 \tiny{$\pm 0.13$} & 96.21 \tiny{$\pm 0.15$}
             \\
\text{ }RF-RGCN            & \textbf{87.62 \tiny{$\pm 0.30$}} & \textbf{83.83 \tiny{$\pm 0.52$}} & \textbf{83.92 \tiny{$\pm 0.21$}}
                   & \textbf{83.37 \tiny{$\pm 0.27$}} & \textbf{96.74 \tiny{$\pm 0.09$}} & \textbf{96.47 \tiny{$\pm 0.10$}}
             \\
\cline{1-7}
\noalign{\vskip\doublerulesep
         \vskip-\arrayrulewidth} \cline{1-7}
\rule{0pt}{2.5ex}
RGCN               & 86.80 \tiny{$\pm 0.27$} & 82.72 \tiny{$\pm 0.27$} & 80.42 \tiny{$\pm 0.73$}
                   & 80.10 \tiny{$\pm 0.25$} & 96.52 \tiny{$\pm 0.32$} & 96.24 \tiny{$\pm 0.36$}
             \\
\cdashline{1-7}
\rule{0pt}{2.5ex}
RGAT               & 86.93 \tiny{$\pm 0.35$} & 82.83 \tiny{$\pm 0.32$} & 80.44 \tiny{$\pm 0.82$}
                   & 80.12 \tiny{$\pm 0.33$} & 96.52 \tiny{$\pm 0.28$} & 96.24 \tiny{$\pm 0.25$}
             \\
\text{ }RF-RGAT-\textit{E} & 87.32 \tiny{$\pm 0.31$} & 83.24 \tiny{$\pm 0.38$} & 82.45 \tiny{$\pm 0.72$}
                   & 82.08 \tiny{$\pm 0.65$} & 96.76 \tiny{$\pm 0.13$} & 96.54 \tiny{$\pm 0.12$}
             \\
\text{ }RF-RGAT            & \textbf{87.86 \tiny{$\pm 0.24$}} & \textbf{83.99 \tiny{$\pm 0.10$}} & \textbf{83.96 \tiny{$\pm 0.32$}}
                   & \textbf{83.45 \tiny{$\pm 0.28$}} & \textbf{96.76 \tiny{$\pm 0.13$}} & \textbf{96.54 \tiny{$\pm 0.12$}}
             \\
\bottomrule
\end{tabular}
\end{adjustbox}
\end{scriptsize}
\end{center}
\label{tb:extensibility}
\end{table*}

When using a heterogeneous GNNs, $\mathcal{E}=\left\{\mathcal{E}^{1}, \mathcal{E}^{2}\right\}$, the adjacency matrices $\boldsymbol{A}^{1}$ and $\boldsymbol{A}^{2}$ represent the followers and friends relationships, respectively. By leveraging more information from user relationships, the heterogeneous GNNs performs better than the homogeneous GNNs. The results in Table \ref{tb:extensibility} demonstrate the effectiveness of the three modules in RF-GNN. On all datasets, our proposed RF-GNN applied to the heterogeneous GNNs consistently enhances the model's performance.

\subsection{Robustness of the model}
\label{sec:Robustness}
\begin{figure}[h]
  \centering
   \begin{minipage}{0.965\textwidth}
    \centerline{\includegraphics[width=\textwidth]{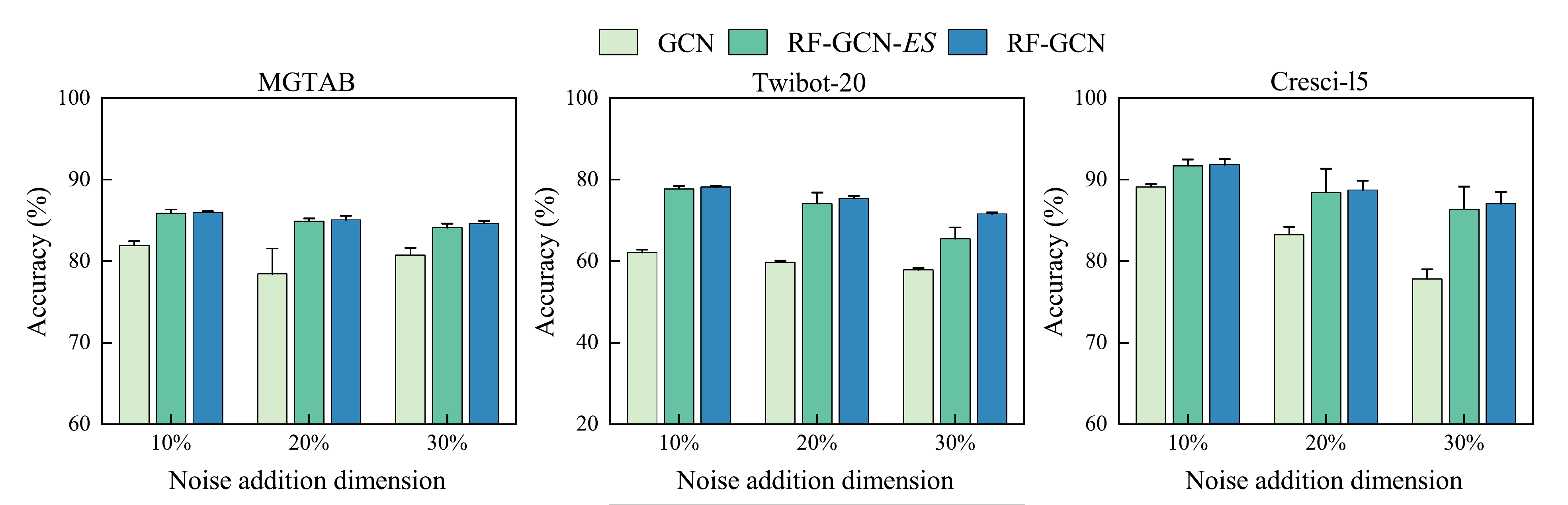}}
   \end{minipage}
  \caption{The performance of RF-GCN with different proportion of random noise.}
  \label{fig:random_noise}
\end{figure}

In this section, we evaluate the performance of various model under noise to verify the robustness of the models. In social bot detection, bots may change their registration information to evade detection. We simulate this scenario by adding a certain proportion of random noise to the feature vectors of the accounts. Specifically, we randomly add Gaussian noise with a mean of 0 and a variance of 1 to 10\%, 20\%, and 30\% of the features. The dataset partition and parameter settings are the same as in Section \ref{sec:Overall_performance}, and the experimental results are shown in Figure \ref{fig:random_noise}.

The RF-GCN-\textit{ES} removes the alignment mechanism compared to the RF-GCN model. The results in Figure \ref{fig:random_noise} show that RF-GCN performs better than RF-GCN-\textit{ES} and GCN on all datasets under different noise ratios, demonstrating that the alignment mechanism proposed in our framework improves the accuracy of the model. As the noise ratio increases, the decrease in accuracy for RF-GCN is less than that of RF-GCN-\textit{ES} and GCN, demonstrating that our proposed alignment mechanism increases the robustness of the model.

\section{Related work}
\subsection{Social bot detection}
Socail bot detection methods can be broadly divided into two categories: feature-based and graph-based approaches. Feature-based methods rely on feature engineering to design or extract effective detection features, and then use machine learning classifiers for classification. In earlier studies~\cite{article03,article32}, features such as the number of followers and friends, the number of tweets, and the creation date were used. Other studies have incorporated features derived from user tweets~\cite{article32,article33}. To identify social bots using extracted user features, machine learning algorithms have been employed in numerous studies~\cite{article01,article02,article03}. In particular, ensemble learning techniques such as Random Forest~\cite{article04}, Adaboost~\cite{article08}, XGBoost~\cite{article28} have been applied in bot detection. Although feature-based methods are simple and efficient, they fail to utilize the interaction relationships between users. Graph neural network-based machine account detection methods~\cite{article13,article15,article16} convert account detection into a graph node classification problem by constructing a social relationship graph of users. Compared to feature-based methods, graph-based methods have increasingly gained attention for their effective use of user interaction features such as follow and friend relationships~\cite{article12}.

\subsection{Semi-supervised Learning on Graphs}
DeepWalk~\cite{article26} is the first work for graph embedding. As an unsupervised method for learning latent representations of nodes, DeepWalk can easily be turned into a semi-supervised baseline model when combined with an SVM classifier. Compared to DeepWalk's random walk strategy for graph embedding, Node2Vec~\cite{article27} balances homogeneity and structural results by adjusting the probability of random walks.
Graph neural network (GNN) models have achieved enormous success. Originally inspired by graph spectral theory,~\cite{article24} first designed a learnable graph convolution operation in the Fourier domain. The model proposed by~\cite{article17} simplifies the convolution operation by using a linear filter and has become the most prevailing one. GAT~\cite{article19} proposed to use an attention mechanism to weight the feature sum of neighboring nodes on the basis of GCN. Many algorithms~\cite{article21,article22,article23} have improved GCN and boosted the performance of graph neural networks.
Recently, some methods have combined ensemble learning with graph neural networks and achieved good results. In graph classification tasks, XGraphBoost~\cite{article25} first formats the original molecular data into a graph structure, then uses a graph neural network to learn the graph representation of molecular features, and finally loads the graph representation as sample features into the supervised learning model XGBoost. XGraphBoost can facilitate efficient and accurate prediction of various molecular properties. In node classification tasks, Boosting-GNN~\cite{article20} combines GNN with the Adaboost algorithm to propose a graph representation learning framework that improves node classification performance under unbalanced conditions.

\section{Conclusion}
In this paper, we proposed a novel technique for detecting social bots, Random Forest boosted Graph Neural Network (RF-GNN), which employs GNN as the base classifier for the random forest. Through subgraph construction, a set of subgraphs are generated, and different subgraphs are used to train the GNN base classifiers. The outputs of different branches are then integrated. Additionally, we propose the aligning mechanism, which leverages the output of GNNs and FCNs, to further enhance performance and robustness. Our proposed method effectively harnesses the advantages of GNNs and ensemble learning, resulting in a significant improvement in the detection performance and robustness of GNN models. Our experiments demonstrate that RF-GNN consistently outperform the state-of-the-art GNN baselines on bot detection benchmark datasets.

\section*{CRediT authorship contribution statement}
\textbf{Shuhao Shi:} Conceptualization, Investigation, Methodology, Software, Validation, Analysis, Writing, Visualization. \textbf{Qiao Kai:} Conceptualization, Investigation, Methodology, Software, Validation, Analysis,  Writing, Visualization. \textbf{Jie Yang:} Methodology, Validation, Writing, Supervision. \textbf{Baojie Song:} Methodology, Writing, Visualization. \textbf{Jian Chen:} Conceptualization, Methodology, Supervision. \textbf{Bin Yan:} Conceptualization, Methodology. Supervision.

\section*{Declaration of competing interest}
The authors declare that they have no known competing financial interests or personal relationships that could have appeared to influence the work reported in this paper.

\section*{Data availability}
The data used in this paper are all public datasets. The source code, data, and other artifacts have been made available at \url{https://github.com/GraphDetec/RF-GNN}.

\section*{Acknowledgment}
This work was supported by the National Key Research and Development Project of China (Grant No. 2020YFC1522002).


\bibliographystyle{elsarticle-num}
\bibliography{ecrc}






% \end{thebibliography}

\end{document}